%% file: main.tex
\documentclass[10pt]{article} 
\usepackage[accepted]{tmlr}

\input{math_commands.tex}

\usepackage[utf8]{inputenc} 
\usepackage[T1]{fontenc}    
\usepackage{hyperref}       
\hypersetup{
    colorlinks=true,
    citecolor=[rgb]{0.2, 0.4, 1},  
    linkcolor=[rgb]{1., 0.0, 0.0},
    urlcolor=blue
}
\usepackage{url}            
\usepackage{booktabs}       
\usepackage{amsfonts}       
\usepackage{nicefrac}       
\usepackage{microtype}      

\usepackage[table,xcdraw]{xcolor}
\usepackage[normalem]{ulem}
\useunder{\uline}{\ul}{}
\usepackage{graphicx}
\usepackage{subfigure}
\usepackage{wrapfig}
\usepackage{multirow}
\usepackage{pgfplots}
\usepackage{natbib}
 \usepackage{amsmath}
 \usepackage{cleveref}
 \usepackage{adjustbox}
\usepackage{tcolorbox}
\usepackage{tabularx}
\usepackage[framemethod=TikZ]{mdframed}

\definecolor{darkgreen}{rgb}{0.27, 0.66, 0.44}
\definecolor{lightblue}{rgb}{0.25, 0.5, 0.85}
\definecolor{lightgray}{gray}{0.85}
\definecolor{seabornblue}{RGB}{ 31, 119, 180}
\definecolor{darkred}{RGB}{139, 0, 0}

\crefname{figure}{Fig.}{Figs.}
\Crefname{figure}{Fig.}{Figs.}
\crefname{table}{Tab.}{Tabs.}
\Crefname{table}{Tab.}{Tabs.}
\crefname{section}{Sec.}{Secs.}
\Crefname{section}{Sec.}{Secs.}

\usepackage{url}

\title{There are no Champions in \\ Supervised Long-Term Time Series Forecasting}

\author{\name Lorenzo Brigato$^{1}$\thanks{Equal contribution. Emails: \{\texttt{name}\}.\{\texttt{lastname}\}\texttt{@unibe.ch}}~,
Rafael Morand$^{1,2,3}$\footnotemark[1]~,
Knut Str\o mmen$^{1,2}$\footnotemark[1]~, \\
Maria Panagiotou$^{1,2}$,
Markus Schmidt$^{3}$,
Stavroula Mougiakakou$^{1}$
\\
\addr
$^{1}$ ARTORG Center, University of Bern, $^{2}$ Graduate School for Cellular and Biomedical Sciences, University of Bern\\
$^{3}$ Center for Experimental Neurology, Department of Neurology, Bern University Hospital
}

\def\month{01}  
\def\year{2026} 
\def\openreview{\url{https://openreview.net/forum?id=yO1JuBpTBB}} 

\begin{document}

\maketitle

\begin{abstract}
Recent advances in long-term time series forecasting have introduced numerous complex supervised prediction models that consistently outperform previously published architectures.
However, this rapid progression raises concerns regarding inconsistent benchmarking and reporting practices, which may undermine the reliability of these comparisons.
In this study, we first perform a broad, thorough, and reproducible evaluation of the top-performing supervised models on the most popular benchmark and additional baselines representing the most active architecture families. This extensive evaluation assesses eight models on 14 datasets, encompassing $\sim$5,000 trained networks for the hyperparameter (HP) searches.
Then, through a comprehensive analysis, we find that slight changes to experimental setups or current evaluation metrics drastically shift the common belief that newly published results are advancing the state of the art.
Our findings emphasize the need to shift focus away from pursuing ever-more complex models, towards enhancing benchmarking practices through rigorous and standardized evaluations that enable more substantiated claims, including reproducible HP setups and statistical testing.
We offer recommendations for future research.

\end{abstract}

\section{Introduction}


\begin{wrapfigure}{r}{0.42\textwidth}  
    \vspace{-0.3cm}
    \centering
    \centerline{\includegraphics[width=0.42\columnwidth]{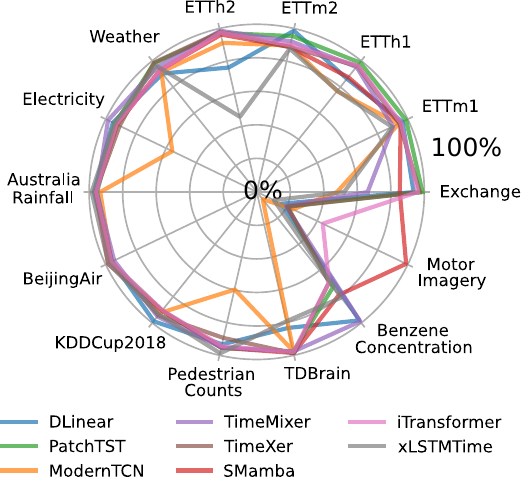}}
    \caption{\textbf{There is no champion.}
        The relative MSE averaged over all forecast horizons reveals that no model dominates on all datasets.}
        \label{fig:radarplot}
        \vspace{-1cm}
\end{wrapfigure}

Long-term time series forecasting (LTSF) is critical across various domains, including energy management \citep{WERON20141030}, financial planning \citep{SEZER2020106181}, and environmental modeling \citep{10825393}. 
Accurately predicting future values in time series data enables better decision-making and resource allocation. 
LTSF remains challenging due to the complex temporal dynamics, including trends, seasonality, irregular fluctuations, and significant variability across datasets \citep{qiu2024tfb,shao2024exploring}.

Recent advances in deep learning have improved LTSF capabilities, and the field is currently witnessing an exponential surge in publication rates \citep{kim2024comprehensive}.
Popular research directions within the field include supervised LTSF, which involves training and testing models on IID data from historical time series \citep{wang_timemixer_2024, liu_itransformer_2024, wang2025mamba}, in contrast to the pre-training and zero-shot or fine-tuning paradigms introduced by foundation models \citep{cao2023tempo, woo2024unified}.
From an architectural perspective, transformer models have been adapted to time series forecasting with innovative modifications, such as univariate patching \citep{nie_time_2023} and attention mechanisms tailored to exploit inter-variate dependencies \citep{liu_itransformer_2024,wang_timexer_2024}.
In addition, models leveraging multiscale signal mixing \citep{wang_timemixer_2024}, Fourier-based 2D decomposition \citep{wu_timesnet_2023}, state-space modeling \citep{wang2025mamba}, 1D convolution \citep{luo2024moderntcn}, and novel recurrent processing \citep{beck2024xlstm, alharthi2024xlstmtime} have expanded the field.

However, we claim that the field is facing significant challenges regarding fair benchmarking, transparent reporting, and guidelines for model selection. 
Although this problem is not uniquely encountered in LTSF \citep{eriksson2025_benchmarking,herrmann2024position}, we highlight the specific challenges in LTSF to promote discussions towards improvements in the field, as it was done in other disciplines \citep{bechler2025_graphbenchmarking,mcintosh2024_llmbenchmarking,sarfraz24a_quo_vadis,wu_current_2023}.
In this work, we focus on supervised LTSF and therefore exclude foundation models, as their evaluation requires distinct experimental protocols involving large-scale pre-training with potential data leakage concerns \citep{aksu2024gift}.
We observe inconsistencies in test setups across different benchmarks, biased comparisons, and challenges with reproducibility that hinder fair performance assessment in the field.
Moreover, marginal performance gains in recent literature cast doubt on the practical value of increasingly complex model architectures \citep{zeng_are_2022}.
To support our claim, we conduct a comprehensive, rigorous, and reproducible evaluation of the top-performing models on the most widely used benchmark, encompassing five models and three additional baselines representing the most popular neural architectures over 14 datasets ($\sim$5,000 trained networks for the HP searches).
Our results reveal that no single model consistently outperforms all the baselines (\Cref{fig:radarplot}), directly challenging the prevailing narrative of new architectures consistently surpassing competing models across all domains \citep{liu_itransformer_2024,nie_time_2023,wang_timemixer_2024,wang_timexer_2024,wu_timesnet_2023, alharthi2024xlstmtime}.
The findings of our work emphasize the need to shift focus away from pursuing ever-more complex models and towards enhancing benchmarking practices through rigorous and standardized evaluation methods.
We analyze the potential reasons behind this phenomenon and propose recommendations to help the field progress.
To foster reproducibility, our code is available at \url{https://github.com/AIHNlab/NoChamps}.
The contributions of our paper are as follows:

\begin{itemize}
    \item We question the narrative of consistently dominated supervised LTSF benchmarks (\Cref{sec:all_champs}), and support our claim with results obtained through a comprehensive and reproducible experimental setup (\Cref{sec:all_similar}).
    
    \item We challenge previous guidelines for model selection based on dataset characteristics (\Cref{sec:all_similar}) and highlight the need for further research in this direction (\Cref{sec:future_directions}).
    
    \item We investigate potential causes behind overstated claims by carefully analyzing our experimental setup and those from prior work in the literature (\Cref{sec:y_all_champ}), and offer recommendations to help prevent similar issues in future work (\Cref{sec:future_directions}).
\end{itemize}

\section{Field overview}
\label{sec:all_champs}

We provide an overview of recent advancements in LTSF, focusing on current benchmarks (\Cref{sec:recent_bench}) and emerging champions (\Cref{sec:rel_champ}).
Due to space limitations, additional related work on recent time series forecasting models and paradigms is included in \Cref{sec:related_work}.

\subsection{Benchmarks and their recommendations for dataset-guided model selection}
\label{sec:recent_bench}

\textbf{TSLib} \citep{wang_deep_2024} compares 12 deep learning models across five tasks: classification, imputation, anomaly detection, and long-/short-term forecasting. 
For long-term forecasting, nine datasets from four domains are used.
Results are presented for two settings: unified hyperparameters (HPs) and an HP search per model, but details on parameters, context length, forecast horizon, or the search process are missing. 
The evaluation metric is the mean squared error (MSE) averaged across datasets.
The authors claim that their results clearly demonstrate the superior forecasting capabilities of transformer models, particularly iTransformer and PatchTST, despite arguably marginal improvements over MLP-based models, such as N-Beats \citep{Oreshkin2020}. 
They further emphasize continued exploration of temporal-token methods.

\textbf{TFB} \citep{qiu2024tfb} evaluates 22 statistical, classical machine learning, and deep learning methods using 25 multivariate and 8,068 univariate datasets. 
Based on their results, the authors claim that linear models outperform deep learning methods in datasets with increased trends and distribution shifts.
Conversely, transformers excel in datasets with marked patterns (e.g., seasonality).  
Notably, PatchTST and DLinear \citep{zeng_are_2022} consistently perform well across datasets, exhibiting no major weaknesses.

\textbf{BasicTS+} \citep{shao2024exploring} incorporates 28 forecasting models, including 17 short-term forecasting (STF) and 11 LTSF models, across 14 widely used datasets. 
STF models encompass prior-graph-based, latent-graph-based, and non-graph-based methods, while LTSF models consist of transformer-based and linear-layer-based architectures.
Models are implemented following publicly available architectures and HPs, with further tuning of parameters like learning rate and batch size via grid search to ensure performance is at least as good as reported in the original paper.
Upon analyzing the results, the authors argue that dataset characteristics play a major role in determining model performance. 
They claim that transformer models excel on datasets with clear, stable patterns, whereas simpler models like DLinear perform comparably on datasets without such patterns. 
The authors emphasize the need to address data distribution drift and unclear patterns instead of focusing solely on increasing model complexity.
They suggest that this may indicate potential overfitting to commonly used datasets like \textit{ETT*}, \textit{Electricity}, \textit{Weather}, and \textit{Exchange}, which risks creating a misleading impression of progress. 
They conclude that careful dataset selection and curation are essential to advance the field.

\textbf{GIFT-Eval} \citep{aksu2024gift} assesses five statistical forecasting models, eight supervised deep learning models, and four foundation models on 21 LTSF datasets and 55 STF datasets. The implementations of the supervised deep learning models adhere to the original works. This benchmark enables simple assessments of model performances across several dataset characteristics (univariate/multivariate, sampling frequency, domain, and forecast horizon). The authors found that PatchTST offered the most reliable results across all characteristics, while foundation models showed inconsistent performance, suggesting that these models are still in an early and relatively underperforming stage compared to well-tuned supervised approaches, a finding also corroborated by \cite{xu_specialized_2024}.

\subsection{Emergent LTSF champions}
\label{sec:rel_champ}

\begin{wraptable}{r}{0.55\textwidth}
\vspace{-0.8cm}
\renewcommand{\arraystretch}{1.0}
\setlength{\tabcolsep}{12pt}
\caption{\textbf{Model win rates in previous works.} Winners in LTSF for forecast horizons $T \in \mathcal{T} = \{96, 192, 336, 720\}$.
The win rates (\%) are according to each $T$ without averaging.
TimeMixer reported unified parameters (A) and HP search (B) results. 
$\dagger$ avg. over $\mathcal{T}$ and ETT* datasets.}
\label{tab:winners}
\resizebox{\linewidth}{!}{
\begin{tabular}{l|cc}
\toprule
\textbf{Model} & Win \% (MSE) & Win \% (MAE) \\
\midrule
DLinear \citep{zeng_are_2022} & 50.0 & 16.7 \\
\cmidrule{2-3}
PatchTST \citep{nie_time_2023} & 87.5 & 59.4 \\
\cmidrule{2-3}
\multirow{2}{*}{TimeMixer \citep{wang_timemixer_2024}} 
    & A) 93.8 & A) $~$100 \\
    & B) 81.2 & B) 81.2 \\
\cmidrule{2-3}
\multirow{2}{*}{iTransformer \citep{liu_itransformer_2024}} 
    & 33.3 & 47.2 \\
    & 71.4$^\dagger$ & 85.7$^\dagger$ \\
\cmidrule{2-3}
TimeXer \citep{wang_timexer_2024} & 85.7 & 60.7 \\
\bottomrule
\end{tabular}
}
\vspace{-0.5cm}
\end{wraptable}

Recent models have made a leap in LTSF performance \citep{liu_itransformer_2024, nie_time_2023, wang_timemixer_2024, wang_timexer_2024, wu_timesnet_2023, zeng_are_2022}. 
A popular benchmark for supervised LTSF is TSLib \citep{wang_deep_2024}, which, at the time of acceptance, has accumulated over 11.2k stars and 1.8k forks on GitHub, reflecting its wide adoption in the field. 
A series of new models in 2024 has reportedly dominated the field. 
The current leaderboard in TSlib includes five models, originally published in top-tier machine learning conferences.
We summarize the striking win percentages from their original works in \Cref{tab:winners} and briefly describe each model in the following box.

\begin{tcolorbox}[colframe=seabornblue!80, colback=black!4, coltitle=white, colback=white, 
    boxrule=0.5mm, arc=1.5mm, left=2mm, right=2mm, top=1mm, bottom=1mm, 
    fonttitle=\bfseries, title=TSLib Long-Term Time Series Forecasting Leaderboard]
    \footnotesize
        
        We present top TSLib models under fixed and searched look-back settings, ordered by publication year:

        \textbf{DLinear - \citeauthor{zeng_are_2022}} \textit{AAAI-23}: Linear model introduced to challenge transformers in early benchmarks.
        
        \textbf{PatchTST - \citeauthor{nie_time_2023}} \textit{ICLR-23}: Transformer with univariate patching of time series for tokenization.
        
        \textbf{TimeMixer - \citeauthor{wang_timemixer_2024}} \textit{ICLR-24}: MLP-mixer that introduced past-decomposable and future-predictor mixing.
        
        \textbf{iTransformer - \citeauthor{liu_itransformer_2024}} \textit{ICLR-24}: Transformer variant that attends across variates instead of patches.
        
        \textbf{TimeXer - \citeauthor{wang_timexer_2024}} \textit{NeurIPS-24}: Transformer with dual attention for interactions on patches and variates.

\end{tcolorbox}

Beyond these leading models, additional neural architectures have recently gained attention within the community for their focus on efficient computation.
Namely, these approaches include state-space Mamba models \citep{gu2024mamba}, recurrent models via the extended LSTM (xLSTM) architecture \citep{beck2024xlstm}, and fully convolutional networks \citep{luo2024moderntcn}.
We select three models that represent these architecture families, were explicitly adapted or designed for LTSF, and gained rapid interest within the community (measured in terms of citations). Specifically, we include S-Mamba \citep{wang2025mamba}, xLSTMTime \citep{alharthi2024xlstmtime}, and ModernTCN \citep{luo2024moderntcn}.

\section{Who is the real champion?}
\label{sec:all_similar}

\textbf{Motivating \textit{confirmatory research} in LTSF.} As seen in the previous section, recent papers often suggest that newly proposed architectures outperform others across almost all tested datasets. 
However, the variability in results for the same algorithms and reliance on prior studies with different experimental setups raise questions about their consistent superiority. 
This concern goes beyond the field of LTSF and is well supported by a growing body of work criticizing the reliability and generalizability of empirical results in machine learning \citep{bechler2025_graphbenchmarking,eriksson2025_benchmarking,jordan2024position,sarfraz24a_quo_vadis,sculley2018winner,stine2006comment, wu_current_2023}.
In response, the community has increasingly recognized the need for more \textit{confirmatory research} \citep{herrmann2024position}, i.e., empirical evaluations conducted by researchers without vested interest in a particular method, aiming to ensure fairness, minimize bias, and reduce overly optimistic conclusions.
Motivated by the exponential surge in publication rates in time series forecasting \citep{kim2024comprehensive} and the community quest for more neutral benchmarking \citep{herrmann2024position}, we selected the previously introduced five top-performing models from TSLib \citep{wang_deep_2024}, S-Mamba, xLSTMTime, and ModernTCN to perform a comprehensive empirical evaluation across 14 datasets from various domains.
For completeness, a comparison including classical statistical methods is presented in Appendix \ref{sec:full_results}.

\input{table_main_results}

\textbf{Datasets.} We evaluate models on 14 datasets from five domains: Energy, Economy, Transport, Health, and Environment. 
They vary substantially in sample size, sampling frequency, number of variates, and various statistics for time series dynamics (\Cref{sec:dataset_statistics}).
The selection aims to minimize bias by preventing any model from gaining an unfair advantage due to specific dataset characteristics.

\begin{figure}[t]
    \centering
        \centerline{\includegraphics[width=1.0\columnwidth]{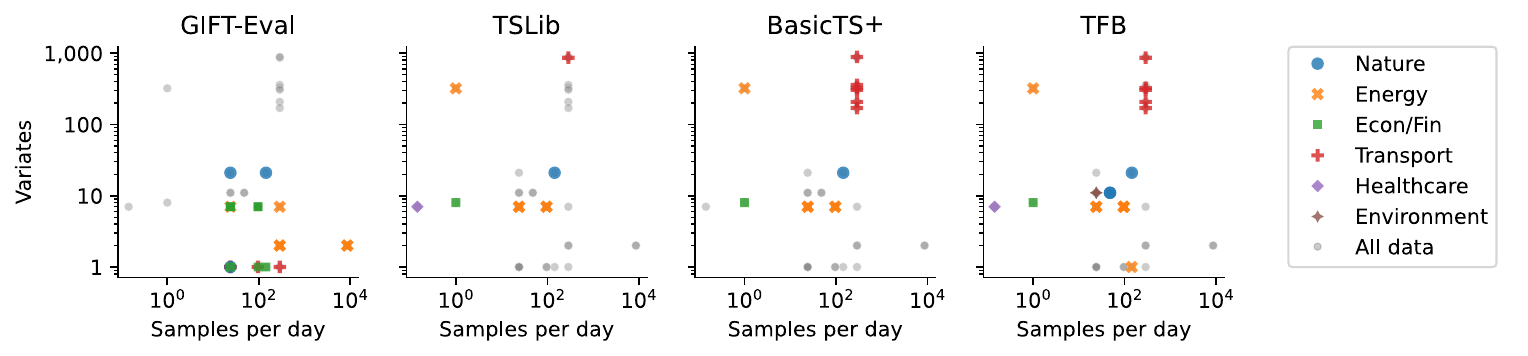}}
        \caption{\textbf{Potential lack in dataset diversity}
        The benchmarks do not span a wide range of frequencies and number of variates across domains.}
        \label{fig:benchmark_dataset_description}
\end{figure}

\textbf{HP search.} We searched HPs aligned with those described in TimeMixer \citep{wang_timemixer_2024}. 
Specifically, we optimized input lengths, learning rates, and the number of encoder layers.
We used the Optuna framework \citep{optuna_2019}, employing the default \texttt{TPEsampler} for HP sampling and the \texttt{SuccessiveHalvingPruner} as the trial scheduler.
We employed Adam with an exponentially decaying scheduler.
For xLSTMTime, we switched the loss function from MSE (default in TSLib) to MAE, following previous implementations \citep{alharthi2024xlstmtime, kraus2024xlstm}, as we indeed observed worse performance using MSE in preliminary experiments.
The optimized HPs were used to train and evaluate the final model across three random seeds to ensure robust results.
For the rest of the model HPs, we default to the configurations provided in TSLib.
We refer the reader to \Cref{sec:hp_search_appendix} for a more detailed description.

\textbf{Finding: There is no champion.} We follow the TSlib benchmark and use MSE and mean absolute error (MAE) to evaluate model performance.
\Cref{tab:main_results} presents the MSE and MAE for each dataset, averaged over the most common forecast horizons \citep{liu_itransformer_2024, nie_time_2023, wang_timexer_2024}, revealing results that differ substantially from recent papers where proposed algorithms often dominate.
Instead, our findings indicate no definitive best-performing model across all datasets and forecast horizons (\Cref{tab:full_results_mean} and \Cref{tab:full_results_min}, \Cref{sec:full_results}).
To assess reliability, we compared our results with the best-reported outcomes from the original studies on three common datasets—\textit{ETT*}, \textit{Electricity}, and \textit{Weather}—and found that our HP search performed similarly or better, confirming the proper implementation and tuning of our baselines (\Cref{tab:setup_reliability}, \Cref{sec:reliability_appendix}).
To present a comprehensive view that highlights both the optimal outcomes and the realistic performance variability, we report the minimum values (best MSE/MAE) and the averages.
We further analyze HP-search run-to-run variability in \Cref{sec:hp_variability}.

\textbf{Finding: Recommendations for dataset-guided model selection do not hold.} We analyze whether our results align with the guidelines proposed by BasicTS+ \citep{shao2024exploring} and TFB \citep{qiu2024tfb} (\Cref{sec:recent_bench}), which aim to address the challenge of selecting appropriate models for specific datasets in LTSF.

\begin{wrapfigure}{r}{0.55\textwidth} 
    \begin{center}
    \vspace{-0.5cm}
        \centerline{\includegraphics[width=0.55\columnwidth]{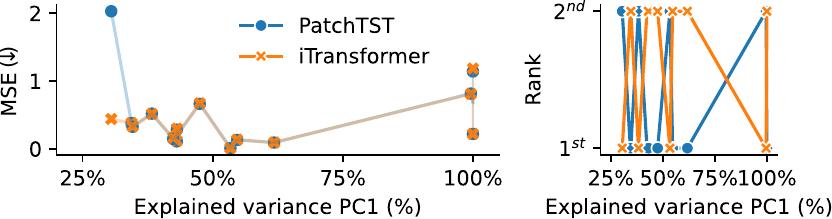}}
        \caption{\textbf{Uni- vs. multivariate}
        PatchTST and iTransformer perform comparably in terms of explained variance.}
        \label{fig:comp_variate_models}
        \vspace{-0.75cm}
    \end{center}
\end{wrapfigure}

We revisit the example in \citep{shao2024exploring} and assess the performance of a linear model (DLinear) versus a transformer model (PatchTST) on data with clear and unclear patterns, respectively. 
We use \textit{Exchange} as the dataset with an unclear pattern and \textit{PedestrianCounts} as the dataset with a clear pattern (\Cref{fig:data_patterns}, \Cref{sec:clear_vs_unclear_ds}).
PatchTST outperforms DLinear on both occasions (\Cref{tab:full_results_mean}, \Cref{sec:full_results}), contradicting the previous recommendations that linear models should be preferred when the data lacks clear patterns  \citep{shao2024exploring}.
Next, to assess guidelines regarding univariate and multivariate data, we compare PatchTST (univariate) against iTransformer (multivariate) on all datasets. 
We use the explained variance from principal component analysis as a proxy for inter-variate similarity (\Cref{sec:dataset_stats}).
Neither model performs increasingly better depending on the inter-variate similarity as ranks fluctuate across the spectrum (\Cref{fig:comp_variate_models}), contradicting the guideline that multivariate models should be preferred if the data has strong inter-variate similarities \citep{qiu2024tfb,shao2024exploring}. 

Furthermore, GIFT-Eval claims to provide insights into the strengths of different models across domains, frequencies, prediction lengths, and the number of variates. 
To investigate this, we compiled all datasets included in the LTSF benchmarks and analyzed their distribution by sampling frequency, number of variates, and domain, revealing a substantial imbalance (\Cref{fig:benchmark_dataset_description}).
For instance, GIFT-Eval represents the \textit{Transport} domain with two univariate datasets, while BasicTS+ and TFB include several multivariate transport datasets, which is more consistent with the graph-structured nature of such systems \citep{shao2024exploring, qiu2024tfb}. 
Moreover, no benchmark includes high-frequency datasets (e.g., EEG or wearable-sensor data), excluding many real-world applications.
Hence, we speculate that LTSF benchmarks lack sufficient dataset variety to enable systematic, dataset-guided model evaluation, ultimately hampering progress toward understanding when and why particular models succeed in LTSF.

\section{Why are they all champions?}
\label{sec:y_all_champ}

\begin{figure*}[t]
    \centering
    \adjustbox{valign=c, width=0.35\linewidth}{\input{plots/system_exclude_motor.pgf}}
    \hfill
    \adjustbox{valign=c, width=0.35\linewidth}{\input{plots/system_exclude_nomotor.pgf}}
    \hfill
    \adjustbox{valign=c, width=0.25\linewidth}{
        \begin{tabular}{lcc}
            \toprule
            W/ \textit{MotorImagery} & Win \% & MSE \\
            \midrule
            S-Mamba        & \textbf{\textcolor{red}{55.68}} & \textbf{\textcolor{red}{0.49 $\pm$ 0.16}} \\
            PatchTST      & \textcolor{blue}{\underline{17.58}} & 0.69 $\pm$ 0.38 \\
            TimeXer       & 10.58 & 0.68 $\pm$ 0.38 \\
            DLinear       & 8.18  & 0.77 $\pm$ 0.46 \\
            TimeMixer     & 5.10  & 0.71 $\pm$ 0.39 \\
            xLSTMTime     & 2.56  & 0.97 $\pm$ 0.64 \\
            iTransformer  & 0.28  & \underline{\textcolor{blue}{0.55 $\pm$ 0.21}} \\
            ModernTCN     & 0.04  & 0.72 $\pm$ 0.31 \\
            \midrule
            W/o \textit{MotorImagery} & Win \% & MSE \\
            \midrule
            PatchTST      & \textbf{\textcolor{red}{46.78}} & \textbf{\textcolor{red}{0.45 $\pm$ 0.14}} \\
            TimeXer       & \textcolor{blue}{\underline{24.48}} & \textbf{\textcolor{red}{0.45 $\pm$ 0.14}} \\
            TimeMixer     & 9.46 & 0.47 $\pm$ 0.14 \\
            DLinear       & 11.84 & 0.48 $\pm$ 0.15 \\
            iTransformer  & 0.96 & \textcolor{blue}{\underline{0.46 $\pm$ 0.14}} \\
            S-Mamba        & 2.70 & \textcolor{blue}{\underline{0.46 $\pm$ 0.14}} \\
            xLSTMTime     & 3.70 & 0.54 $\pm$ 0.17 \\
            ModernTCN     & 0.08 & 0.53 $\pm$ 0.15 \\
            \bottomrule
        \end{tabular}
    }
    \caption{\textbf{Model rankings are highly sensitive to dataset and horizon selection.} We assess the robustness of rankings across 5,000 experimental configurations, each using a random subset of datasets and forecast horizons.
    Including \textit{MotorImagery}, the only dataset with clear model gaps (\Cref{fig:all_ds_violin}, \Cref{sec:full_results}), favors S-Mamba, while excluding it yields close performance across models.
    This highlights the brittleness of current benchmarks, where small changes in datasets or forecast horizons can easily shift which model appears as a champion.
    \textbf{\textcolor{red}{Best}} and {\textcolor{blue}{\underline{second-best}}} are highlighted.}
    \label{fig:exclude_ds_horizon}
\end{figure*}

\subsection{Impact of selective inclusion of datasets and forecast horizons} 
\label{sec:sel_inclusion_ds_horizon}

\textbf{Model rankings are highly sensitive to dataset and horizon selection.} We systematically analyze how the selective exclusion of datasets and forecast horizons in the experimental settings may affect overall rankings.
Let \(\mathcal{D} = \{D_1, \dots, D_M\}\) be the collection of datasets and \(\mathcal{T} = \{T_1, \dots, T_H\}\) the forecast horizons.
We sample \(K = 5,\!000\) experimental configurations, each defined by uniformly drawn subsets of datasets and horizons: \(\mathcal{S}_D \subseteq \mathcal{D},\ |\mathcal{S}_D| = k,\ k \sim \mathcal{U}\{1, M\}\), and \(\mathcal{S}_T \subseteq \mathcal{T},\ |\mathcal{S}_T| = \ell,\ \ell \sim \mathcal{U}\{1, H\}\).
When all datasets are included, S-Mamba seems to clearly be the best model, obtaining a win percentage of $\sim$56\% and the lowest average MSE of 0.49 over the distribution (\Cref{fig:exclude_ds_horizon}, left).
However, after repeating the analysis without \textit{MotorImagery}, i.e., the only dataset with a clear performance gap among baselines (\Cref{fig:all_ds_violin}, \Cref{sec:full_results}), the distributions overlap (\Cref{fig:exclude_ds_horizon}, center), implying equivalency of all models.
Since win percentages and average MSEs are similar across models, minor changes in datasets and forecast horizons can shift which model appears best, supporting our view regarding the brittleness of current model superiority claims. 

\textbf{Prior work employed inconsistent dataset and horizon selection.} We observe that subsets of the full benchmark were occasionally used, which may be justified, e.g., for lack of dataset size. 
In iTransformer \citep{liu_itransformer_2024}, the authors averaged the performance over the ETT datasets, which they justified by their intrinsic similarity.
However, this increased the percentage of wins of their proposed method from $33.3\%$ to $71.4\%$. BasicTS+
\citep{shao2024exploring} focuses solely on a forecast horizon of 336, whereas TFB \citep{qiu2024tfb} bases the analysis of the impact of different data characteristics on a horizon of 96, despite reporting performance for all four forecast horizons.

\subsection{Impact of selective inclusion of baseline models} 
\label{sec:baseline_inclusion}

\textbf{Model exclusion reshape leaderboards.} While unsurprising, we stress that excluding the top model from a benchmark may automatically crown the second-best as a champion.
For instance, removing S-Mamba from \Cref{tab:main_results} would champion iTransformer as it scores the second-best average metrics. 

\textbf{Prior work overlooked SOTA models in the analyses.} We notice the lack of inclusion of baselines like N-Beats in the publications of recent ``champions'', although it is a top-3 method in \citep{wang_deep_2024} (\Cref{sec:baselines}).
In addition, we identified cases where the best-performing model was excluded from discussions without justification.
For example, \citep{qiu2024tfb,shao2024exploring} claim recent transformers underperform compared to earlier methods, but their experiments show the most recent LTSF transformer at the time (PatchTST) outperformed competitors, contradicting this claim.
Moreover, \citep{shao2024exploring} claims linear models are better for LTSF on datasets with unclear patterns or distribution shifts. However, the claim is based on experiments that excluded PatchTST, thereby weakening the strength of their conclusion.

\subsection{Impact of hyperparameters} 
\label{sec:HP_impact}

\textbf{HP search raises absolute performance.} The impact of HP tuning on the benchmarking performance of models is becoming increasingly evident \citep{Brigato_2021_ICCV, mcelfresh2024neural}.
We investigate whether this is also the case in LTSF via a proof-of-concept example.
We report the evaluation in terms of MSE for DLinear on the \textit{Weather} dataset at forecast horizon 96 (\Cref{tab:impact_hp_tuning}).
HP tuning brings a relative performance boost of $\sim$15\% in our setup and an $\sim$10\% in \citep{wang_deep_2024}.
Similarly, building on the HP search details in \citep{wang_timexer_2024}, we found comparable performance between TimeMixer, iTransformer, and PatchTST, unlike the original work, where TimeMixer consistently ranks first.
This underscores how close the actual performance of models is, making outcomes and conclusions sensitive to slight variations in HP search.

\textbf{Prior work put models at disadvantage through unified HP setup.} In TSLib, models are usually based on the implementation of the original publications.
However, in \citep{wu_timesnet_2023}, for a fair comparison, they changed the input embeddings and the final projections to be the same for all models. 
Specifically, the sequence length was set to 96 for all models by default.
This is critical since DLinear \citep{zeng_are_2022}, after a broad ablation study, explicitly states that short input sequences ($<336$) lead to underfitting.

\begin{table*}[t]
\renewcommand{\arraystretch}{1.5}
\setlength{\tabcolsep}{2pt}
\caption{\textbf{HP search sensitivity.} We report the MSE of DLinear for \textit{Weather} at prediction length 96 when HP tuning is/is not performed, both in our and previous papers, along with the \textbf{\textcolor{darkgreen}{relative performance improvement}} expressed in \% (when possible).
``$-$'' indicates a missing analysis. 
}
\label{tab:impact_hp_tuning}
\centering
\small
\resizebox{\linewidth}{!}{
\begin{tabular}{l|ccccccc}
\toprule
\textbf{DLinear (MSE)} & \citep{zeng_are_2022} & \citep{nie_time_2023} & \citep{wu_timesnet_2023} & \citep{liu_itransformer_2024} & \citep{wang_timemixer_2024} & \citep{wang_timexer_2024} & Ours \\
\midrule
Unified HP & -- & -- & 0.196 & 0.196 & 0.195 & 0.196 & 0.198 \\
HP tuning  & 0.176 & 0.176 & -- & -- & 0.176 & -- & 0.168 \\
Rel. Improv. & -- & -- & -- & -- & \textbf{\textcolor{darkgreen}{+9.7\%}} & -- & \textbf{\textcolor{darkgreen}{+15.1\%}} \\
\bottomrule
\end{tabular}
}
\end{table*}

\subsection{Influence of visualizations}
\label{sec:visualizations}


\begin{figure}[t]
    \centering
    \includegraphics[width=0.7\columnwidth]{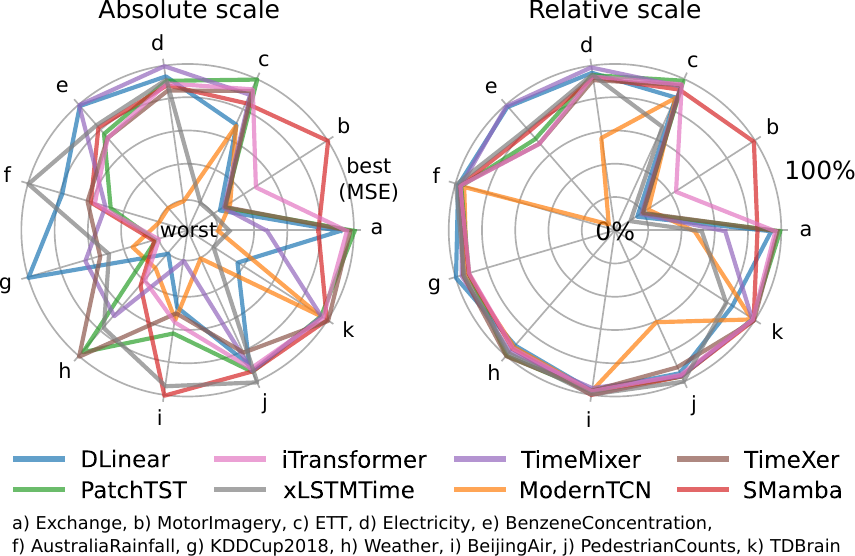}
    \caption{\textbf{Bias in visualizations.}
    The plots show the same results (MSE) represented at two scales.
    The relative scale makes performance differences between models appear more subtle.}
    \label{fig:bias_in_visualizations}
\end{figure}

\textbf{Visualizations may bias perceived rankings.} Visualizations are a strong tool to convey a message. 
In \Cref{fig:bias_in_visualizations}, we investigate the impact of scales to visualize results.
We observe that using an absolute scale in radar plots exaggerates differences between models that are not perceived in the relative scale with uniform axes. 
Conversely, it can obscure substantial differences, as in the example of \textit{MotorImagery} and \textit{BenzeneConcentration} (\Cref{fig:bias_in_visualizations}, right, b and e). 

\textbf{Prior work employed misleading visualization practices.} We identified two examples that used radar plots with absolute scales to visualize the model performances between models \citep{liu_itransformer_2024} and across dataset characteristics \citep{qiu2024tfb}, respectively.
These choices can create a misleading impression of the models’ actual performance and lead to false conclusions. 
Note that other plots may also create biased impressions through axis scaling or selective metric representation.

\subsection{Statistical evidence for model superiority}
\label{sec:statistical_analysis}

\begin{wraptable}{r}{0.55\textwidth}
\vspace{-0.8cm}
\renewcommand{\arraystretch}{1.6}
\setlength{\tabcolsep}{12pt}
\caption{\textbf{iPatch as a questionable champion.}
Although iPatch scores the best average rank and the third-best MSE/MAE averaged over all datasets, it does not statistically differ from all the other baselines under a Friedman test.
\textbf{\textcolor{red}{Best}} and {\textcolor{blue}{\underline{second-best}}} are highlighted.
}
\vspace{0.1cm}
\resizebox{\linewidth}{!}{
\begin{tabular}{l|cccc}
\toprule
Model & MSE & MAE & Rank $\raisebox{0.1ex}{\scriptsize{\text{(MSE)}}}$ & Rank $\raisebox{0.1ex}{\scriptsize{\text{(MAE)}}}$ \\
\midrule
DLinear & 0.770 & 0.484 & 4.93 & 5.00 \\
PatchTST & 0.691 & 0.458 & \textcolor{blue}{\underline{4.07}} & \textcolor{blue}{\underline{4.50}} \\
iTransformer & \textcolor{blue}{\underline{0.547}} & \textcolor{blue}{\underline{0.416}} & 5.07 & 5.29 \\
TimeMixer & 0.712 & 0.462 & 4.64 & 4.79 \\
TimeXer & 0.683 & 0.453 & 4.57 & 4.71 \\
iPatch & 0.580 & 0.422 & \textbf{\textcolor{red}{4.00}} & \textbf{\textcolor{red}{4.07}} \\
S-Mamba & \textbf{\textcolor{red}{0.484}} & \textbf{\textcolor{red}{0.407}} & 4.64 & 4.57 \\
xLSTMTime & 0.971 & 0.519 & 5.64 & 5.21 \\
ModernTCN & 0.714 & 0.490 & 7.43 & 6.86 \\
\bottomrule
\end{tabular}

}
\label{tab:ipatch_champ}
\vspace{-1.2cm}
\end{wraptable}

\textbf{Analysis.} We illustrate, through a proof-of-concept example, how the lack of robust statistical testing can lead to false claims regarding models' superiority.
First, we introduce recommended non-parametric statistical tests \citep{demvsar2006statistical}.
Second, we upgrade the existing iTransformer to emulate current model-design proposals and rigorously evaluate it in our setup (see \Cref{sec:all_similar} for details).

\textbf{Statistical tests.}  \citep{demvsar2006statistical} studied various statistical tests for comparing classifiers from both theoretical and empirical perspectives. 
The study recommended a set of simple, reliable, and robust tests for such comparisons.
In particular, the sign test \citep{salzberg1997comparing} compares two classifiers over multiple datasets, and the Friedman test compares various classifiers over multiple datasets (see \Cref{sec:stat_tests}). 

\textbf{iPatch: A proof-of-concept model.} We briefly introduce the iPatch model as a hybrid architecture that integrates design principles of PatchTST into iTransformer, emulating common model design trends observed in recent research.
Let the input series be $\mathbf{x} \in \mathbb{R}^{B \times C \times L}$, where $B$ is the batch size, $C$ is the number of variates, and $L$ is the lookback length.
Firstly, in iPatch, unlike iTransformer, we reshape the input by splitting the sequence into $N$ cycles of length $P$ ($L = N \cdot P$), transforming it from ${B \times C \times L}$ to ${B \times (C \cdot N) \times P}$ to later enable in the layer the temporal attention as in PatchTST.
Each cycle is subsequently embedded to $d_m$, resulting in $\mathbf{z} \in \mathbb{R}^{B \times (C \cdot N) \times d_m}$.  
Secondly, we enhance the attention module as a sequence of two attentions over variates and cycles.
First, the input is reshaped to ${(B \cdot N) \times C \times d_m}$ to isolate the $C$ variates for each cycle and consequently to apply attention over $C$ similarly to iTransformer.
Next, the output of the variate attention is reshaped to ${(B \cdot C) \times N \times d_m}$ to isolate temporal cycles and apply the second attention mechanism similarly to PatchTST.
Since the iTransformer MLPs operating on univariate data were hypothesized to capture time series properties like amplitude and periodicity \citep{liu_itransformer_2024}, we reshape the series to \({(B \cdot N) \times C \times d_m}\) before applying the MLP and finally map it back to ${B \times (C \cdot N) \times d_m}$ for the next transformer layer.
The linear decoder is applied channel-wise to predict \(T\) steps from 
${B \times C \times (d_m \cdot N)}$.

\textbf{Statistical testing substantiates performance gains from targeted architectural adjustments.} Initially, we evaluate the performance of iPatch following either average MSE/MAE \citep{wang_deep_2024} or ranks.
\Cref{tab:ipatch_champ} shows that iPatch achieves the best average rank and the third-best average MSE and MAE.
Complete results for iPatch are available in \Cref{tab:ipatch_results}, \Cref{sec:full_results}.
In line with these outcomes and common practices in the field of LTSF, iPatch may ``almost'' seem the best-performing model.
However, performing the Friedman test, we observe that no model differs from the others considering MAE ($\chi^2_F=9.33$, $p=0.31$).
Given the lack of overall differences, we conduct a focused comparison between iPatch and iTransformer using the sign test \citep{salzberg1997comparing}, since iPatch builds directly upon iTransformer.
Despite their similar architectures, iPatch wins in terms of MSE/MAE on 11 out of 14 datasets, yielding a statistically significant p-value of 0.05 under the sign test (\Cref{sec:stat_tests}), suggesting that targeted architectural modifications can lead to performance improvements.

\textbf{Prior work neglected statistical testing.} The TSLib benchmark \citep{wang_deep_2024} employs averaging for presenting aggregated results.
However, averages are susceptible to outliers \citep{demvsar2006statistical}.
A classifier's strong performance on one dataset can mask weaknesses elsewhere, so we prioritize consistent performance across problems, making dataset averaging unsuitable for evaluation (\Cref{sec:y_all_champ}).
BasicTS+ \citep{shao2024exploring} and TFB \citep{qiu2024tfb} focus on the number of wins achieved by each model but do not incorporate any statistical testing, making conclusions less reliable and harder to communicate in a concise manner.

\subsection{Trade-off between model performance and efficiency}
\label{sec:efficiency_tradeoff}

\textbf{Analysis.} It is also essential to consider other factors that contribute to a model's superiority, such as the trade-offs between performance (e.g., MSE) and efficiency (e.g., training speed or memory consumption) introduced by architectural modifications.

\begin{table}[t]
\vspace{-0.3cm}
\renewcommand{\arraystretch}{1.3}
\setlength{\tabcolsep}{8pt}
\caption{\textbf{Limited gains from increased model complexity.}
Efficiency-weighted performance comparison ($\xi$) relative to DLinear ($\uparrow$ is better).
Despite higher complexity, most models fail to outperform DLinear across multiple performance-weighted efficiency metrics with all datasets (A) and without \textit{MotorImagery} (B).
\texttt{TP} indicates throughput.
\textbf{\textcolor{red}{Best}} and {\textcolor{blue}{\underline{second-best}}} are highlighted.
}
\vspace{0.1cm}
\resizebox{\linewidth}{!}{
\begin{tabular}{l|cccccccccccccccc}
\toprule
\multirow{2}{*}{$\xi(m, \text{DLinear}, \texttt{MSE}, \Phi)$~\raisebox{0.3ex}{\scriptsize{($\uparrow$)}}} 
& \multicolumn{2}{c}{DLinear} 
& \multicolumn{2}{c}{PatchTST} 
& \multicolumn{2}{c}{iTransformer} 
& \multicolumn{2}{c}{TimeMixer} 
& \multicolumn{2}{c}{TimeXer}
& \multicolumn{2}{c}{S-Mamba}
& \multicolumn{2}{c}{xLSTMTime}
& \multicolumn{2}{c}{ModernTCN} \\
& A & B & A & B & A & B & A & B & A & B & A & B & A & B & A & B \\
\midrule
$\Phi = \{\texttt{FLOPs}\}$ 
& 1.00 & \textbf{\textcolor{red}{1.00}} 
& 0.88 & 0.82 
& \textcolor{blue}{\underline{1.21}} & 0.90 
& 0.69 & 0.65 
& 0.78 & 0.72
& \textbf{\textcolor{red}{1.42}} & \textcolor{blue}{\underline{0.92}}
& 0.71 & 0.81
& 0.75 & 0.63 \\

$\Phi = \{\texttt{\#params}\}$ 
& 1.00 & \textbf{\textcolor{red}{1.00}} 
& 0.98 & 0.92 
& \textcolor{blue}{\underline{1.27}} & \textcolor{blue}{\underline{0.94}} 
& 0.95 & 0.89 
& 0.96 & 0.89
& \textbf{\textcolor{red}{1.39}} & 0.91
& 0.70 & 0.79
& 0.74 & 0.62 \\

$\Phi = \{\texttt{TP},\texttt{memory}\}$ \scriptsize{(train)}
& 1.00 & \textbf{\textcolor{red}{1.00}} 
& 0.91 & 0.86 
& \textbf{\textcolor{red}{1.25}} & \textcolor{blue}{\underline{0.92}} 
& 0.83 & 0.79 
& 0.93 & 0.86
& \textcolor{blue}{\underline{1.18}} & 0.77
& 0.67 & 0.75
& 0.77 & 0.64 \\

$\Phi = \{\texttt{TP},\texttt{memory}\}$ \scriptsize{(test)}
& 1.00 & \textbf{\textcolor{red}{1.00}} 
& 0.92 & 0.87 
& \textcolor{blue}{\underline{1.26}} & \textcolor{blue}{\underline{0.93}} 
& 0.88 & 0.83 
& 0.96 & 0.89
& \textbf{\textcolor{red}{1.29}} & 0.83
& 0.68 & 0.76
& 0.78 & 0.66 \\
\bottomrule
\end{tabular}

}
\label{tab:efficiency_tradeoff}
\end{table}

\textbf{Efficiency-weighted error metric.} Drawing inspiration from neural-architecture-search literature \citep{tan2019mnasnet}, we define a composite metric that summarizes prediction quality and efficiency relative to a baseline model.
Let \( m \) denote a candidate model and \( b \) our baseline, with \( \epsilon(\cdot) \) representing our prediction error metric and \( \Phi = \{\phi_k(\cdot)\}_{k=1}^K \) being a set of efficiency metrics.
The efficiency-weighted error metric \(\xi\) is formulated as:
$\xi(m, b, \epsilon, \Phi) = \frac{\epsilon(b)}{\epsilon(m)} \cdot \prod_{k=1}^K   \frac{\phi_k(b)}{\phi_k(m)}^{{s_k w}}$
where \( s_k \in \{-1,+1\} \) controls the metric-specific ratio directionality (lower/higher is better) and \( w \) encodes the relative importance of efficiency.
Thus, the weighted product formulation of $\mathcal{\xi}$ ensures models' error ratios against the baseline are scaled by efficiency ratios weighted by the exponent \(w\).
When \(w=1\), the efficiency ratio scales linearly with the error ratio, while if \(w=0\), the efficiency ratio has zero relevance.
Models with \( \xi > 1 \) outperform the baseline, while \( \xi < 1 \) indicate unfavorable trade-offs.
The baseline always scores a value of one, i.e., when \(m = b\), $\mathcal{\xi} = 1$.
Furthermore, $b$ is chosen such that $0 < \frac{\phi_k(b)}{\phi_k(m)}^{{s_k}} < 1$, ensuring that the efficiency term always penalizes all other models unless compensated by accuracy gains.

\textbf{Performance-efficiency rankings show lack of improvements.} For this analysis, we consider DLinear as the baseline, being the most efficient model across all metrics.
Furthermore, we set $w = 0.07$ so that a 10\% reduction in any efficiency metric corresponds approximately to a 1\% loss in the error-metric ratio $\frac{\epsilon(b)}{\epsilon(m)}$ when the efficiency ratio $\frac{\phi_k(b)}{\phi_k(m)}$ lies in the range $[0.3, 1]$, as $x^{0.07} \approx 0.1x + 0.9$ in this interval.
We select \texttt{FLOPs} and \texttt{\#params} for theoretical complexity, and throughput (\texttt{TP}) with \texttt{memory} usage for practical hardware efficiency.
In \Cref{tab:efficiency_tradeoff}, we report that, with all datasets included (column A), all models except iTransformer and S-Mamba underperform DLinear ($\xi < 1$).
These two models are primarily leading the performance-efficiency leaderboard for their superior results on \textit{MotorImagery}. Indeed, when the \textit{MotorImagery} dataset is excluded (see \Cref{sec:sel_inclusion_ds_horizon}), even iTransformer and S-Mamba no longer achieve the best trade-offs.
These results suggest that the additional architectural complexity does not currently lead to meaningful benefits.
This pattern is clearly captured by efficiency-weighted error metrics such as $\xi$, which summarize trade-offs between accuracy and efficiency across all datasets and horizons.
Further analyses and details on efficiency metric estimation are provided in \Cref{sec:eff_appendix}.

\textbf{Prior work lacks consensus on performance-efficiency trade-offs.} 
Although there are examples that perform certain trade-off analyses \citep{liu_itransformer_2024, wang_timemixer_2024, shao2024exploring, wang_timexer_2024}, their interpretations appear to be limited by selected subsets of datasets, models, and efficiency metrics, lacking a holistic perspective offered by metrics aggregated across all datasets and horizons, thereby conveying incomplete conclusions.

\section{How can the field establish real champions?}
\label{sec:future_directions}

\begin{tcolorbox}[colframe=seabornblue!80, colback=black!4, coltitle=white, colback=white, 
    boxrule=0.5mm, arc=1.5mm, left=2mm, right=2mm, top=1mm, bottom=-2mm, 
    fonttitle=\bfseries, title=Summary of Recommendations]
    \small
    \begin{minipage}[t]{0.25\linewidth}
        \textit{Improving \\ benchmarking practices}
    \end{minipage}%
    \hfill
    \begin{minipage}[t]{0.76\linewidth}
        $\cdot$ Results should be reported across all datasets and forecast horizons (\Cref{sec:sel_inclusion_ds_horizon}). \\
        $\cdot$ Results from the best-performing models should always be included (\Cref{sec:baseline_inclusion}). \\
        $\cdot$ Rigorously tuned HP configurations must be used for all models (\Cref{sec:HP_impact}). \\
        $\cdot$ Third-party evaluations should be encouraged to strengthen reliability. \\
    \end{minipage} 

    \begin{minipage}[t]{0.25\linewidth}
        \textit{Reducing \\ unsubstantiated claims}
    \end{minipage}%
    \hfill
    \begin{minipage}[t]{0.76\linewidth}
        $\cdot$ Visualizations should not exaggerate minor differences (\Cref{sec:visualizations}). \\
        $\cdot$ Statistical tests should be used when comparing models (\Cref{sec:statistical_analysis}).\\
    \end{minipage} 

    \begin{minipage}[t]{0.25\linewidth}
        \textit{Increasing dataset \\diversity and revising \\guidelines for \\model selection}
    \end{minipage}%
    \hfill
    \begin{minipage}[t]{0.76\linewidth}
        $\cdot$ Benchmarks should include datasets that reflect real-world diversity. \\
        $\cdot$ Benchmarks should define forecast horizons informed by dataset characteristics.\\
        $\cdot$ Methodologies relating model-dataset characteristics should be further explored. \\
        $\cdot$ Performance-efficiency trade-offs should be tackled systematically (\Cref{sec:efficiency_tradeoff}).\\

    \end{minipage}
\end{tcolorbox}

Concluding the previous chapters, we provide recommendations that can be tackled by the community. 
To ensure continued progress in LTSF, the field must address persistent shortcomings in benchmarking, evaluation methodology, and guidelines for model selection. 
We outline guidelines aimed at improving transparency, rigor, and the practical relevance of LTSF research.

\textbf{Improving benchmarking practices.} 
Benchmarks should provide rigorously tuned HP configurations for all models, ideally supported by integrated HP optimization tools. 
Benchmark users must report performance consistently across all supported datasets, forecast horizons, and context lengths.
Even minor deviations in experimental setup can dramatically shift performance rankings (\Cref{sec:y_all_champ}), underscoring the need for transparency and standardization.
Additionally, the field would benefit from objective, independent evaluations, in which test sets are withheld and assessed by third parties, e.g., as originally practiced for ImageNet.

\textbf{Reducing unsubstantiated claims.} Researchers should adopt robust statistical testing to supplement performance rankings and mitigate unreliable claims, as exemplified in \Cref{sec:statistical_analysis}. 
Visualizations must be designed with care to avoid distorting perceived differences.
For instance, scale choices can easily exaggerate or obscure performance gaps as highlighted in \Cref{sec:visualizations}.

\textbf{Increasing dataset diversity and revising guidelines for model selection.} To develop effective model selection guidelines, the benchmarks should include datasets to cover a large spectrum of characteristics.
Potential starting points are the UTSD database \citep{liu2024timer} and the LOTSA database \citep{woo2024unified}, as both databases encompass a wide range of datasets with diverse characteristics. 
In addition to providing the data, a crucial step is to define meaningful forecast horizons based on intrinsic dataset characteristics—an issue exemplified by the arbitrary performance on datasets such as \textit{Exchange} \citep{hewamalage_forecast_2023}. 
Then, future studies should focus on datasets where performance varies significantly among SOTA models. 
As illustrated in \Cref{fig:radarplot}, only two datasets—\textit{BenzeneConcentration} and \textit{MotorImagery}—exhibit clearly distinguishable performance patterns, highlighting the need for further investigation into what dataset characteristics drive such differences. 
In this context, we particularly value dedicated studies examining more broadly which architectures succeed or fail under varying conditions, following the style of recent work \citep{chencloser}.  
From a benchmarking perspective, instead, the field should conduct comprehensive evaluations across datasets with diverse characteristics and consistently compare a broad range of model architectures, ensuring that the best-performing architecture for each category, such as linear, MLP, and transformer models, is clearly reported.
Additionally, practical trade-offs between model performance and efficiency should be assessed by systematically analyzing how architectural changes impact computational cost and memory usage (\Cref{sec:efficiency_tradeoff}). Composite metrics such as \(\xi\) can unify performance and efficiency into a single score.
To support such evaluations, benchmarks should provide standardized functionalities for consistent and detailed comparisons.

\section{Discussion and limitations}
\textbf{Objective and scope of our evaluation.} 
While our experimental design enables a broad and reproducible evaluation of recent supervised LTSF models, it also carries inherent limitations.
We focused on a representative subset of recent, high-impact models belonging to the most popular families, i.e., transformers, MLPs, state-space, convolutional, and recurrent models, to capture current evaluation practices in the field.
Our results may be sensitive to specific dataset choices, hyperparameter search ranges, and implementation details, which remain open challenges for reproducible supervised LTSF.
However, our goal was not to establish exhaustive benchmarks or definitive rankings but to show that recent advancements often yield only marginal improvements when evaluated under consistent and controlled settings with experimental variance dominating over architectural advancements.
By emphasizing recent models, we intentionally highlight the present challenges of the field—particularly the difficulty of reliably assessing progress across comparable experimental conditions.
Moreover, while certain models may excel in narrow, context-specific scenarios (e.g., S-Mamba in \textit{MotorImagery}), such isolated successes do not translate into universal applicability, further supporting our argument against the “champion”
narrative.

\textbf{Setting of fixed ``long-term'' forecast horizons.} A limitation of our study lies in the use of fixed forecast horizons across datasets from different domains which can render the notion of ``long-term'' arbitrary and detached from domain-specific constraints.
Although not necessarily optimal, the chosen forecast horizons reflect current practice and enable comparability with past work.
In particular, we adopt the horizons used in TSLib, which are consistent with those in other recent benchmarks, such as BasicTS+ and TFB, further aligning with the ranges reported in the original publications.
Identifying truly meaningful forecast horizons remains an open challenge, with recent efforts aiming to define horizons in a more data-informed manner \citep{aksu2024gift}, complicated by the unclear distinction between short- and long-term forecasting.
Our findings may also be applicable to shorter horizons, although this requires empirical testing.

\textbf{Exclusion of foundation models.}
While recent trends in time series research increasingly explore the development of foundation models \citep{ShiMML24, yao2025towards}, including multimodal large language models (LLMs) \citep{jin_position_2024}, as SOTA time series forecasters, we purposely excluded their evaluation from our work.
Practically, substantial differences in terms of benchmarking compared to supervised models hold, including factors such as potential data leakage and the considerable computational cost of pre-training \citep{aksu2024gift}.
Dedicated benchmarks are more suitable for critically evaluating and moving forward this parallel line of research in LTSF \citep{aksu2024gift}.
In this regard, the claims following our evaluations do not directly apply to this set of models, given the lack of empirical evidence.
However, we argue that incorporating them would be unlikely to alter the conclusions of our work 
considering recent studies questioning their actual effectiveness \citep{xu_specialized_2024, aksu2024gift, tan_are_2024, bergmeir2024llms, karaouli2025foundational, less-is-more-prune-then-finetune}.
Specifically, in GIFT-Eval, the best-performing foundation model, MOIRAI \citep{woo2024unified}, did not outperform PatchTST on medium and long-term forecasts, even without any HP optimization applied to the latter.
This result highlights that these models are still in an early and relatively underperforming stage compared to well-tuned supervised baselines, a finding also corroborated by \citep{xu_specialized_2024}.
Therefore, while we concur that future work should revisit this question as the field progresses, some of the insights derived from our study may prove valuable for future benchmarking efforts involving foundation models (e.g., rigorous statistical testing).
To acknowledge the growing interest in this direction, we include a brief overview of recent developments in foundation models for time series in \Cref{sec:foundational_models}, as well as another description of LLM-based approaches in \Cref{sec:llms}.

\section{Conclusions}

In this work, we critically evaluate supervised LTSF research and put forward a proposal to address persistent issues in the field.
Importantly, our aim is not to criticize prior work in this longstanding and recently revitalized domain but to provide a constructive analysis that supports both our own work and future research in the field, including its translation into domain-specific applications.
Through an extensive and reproducible evaluation of eight models across 14 datasets, we demonstrated that claims of consistent performance improvements in newly published models often rely on specific experimental setups and evaluation methods.
Our findings question the idea of universal advancements, revealing that no single model consistently excels across our experiments.
We identified issues in the supervised LTSF domain, such as non-standardized evaluation frameworks, biased comparisons, and limited reproducibility that hinder fair assessment and delay real progress.
To address these challenges, we propose a set of actionable recommendations: adopt standardized evaluation protocols, prioritize benchmarking robustness over architectural complexity, and deepen the analysis of how dataset characteristics influence model performance.

\paragraph{Acknowledgments.}
We would like to thank the anonymous reviewers for their constructive feedback, which has improved our manuscript.
This work was partly supported by the European Commission and the Swiss Confederation - State Secretariat for Education, Research
and Innovation (SERI) within the project 101057730 Mobile Artificial Intelligence Solution for Diabetes Adaptive Care (MELISSA) and by the Stiftung Sanitas within the framework of the Sanitas Diabetes Technologie 2.0 project.

\bibliography{main}
\bibliographystyle{tmlr}

\newpage
\appendix

\section{Related work}
\label{sec:related_work}

\subsection{Classical approaches}
Traditional statistical methods, such as AutoRegressive Integrated Moving Average \citep{BoxPierce1970}, Vector Autoregression \citep{toda1993}, Exponential Smoothing \citep{hyndman2008forecasting}, and Spectral Analysis \citep{koopmans1995spectral} were widely used in TS forecasting. 
Progressively, machine learning models such as XGBoost \citep{chen2016xgboost}, Random Forest \citep{breiman2001random}, Gradient Boosting Regression Trees \citep{friedman2001greedy}, and LightGBM \citep{ke2017lightgbm} have shown improvements in the forecast due to their ability to handle non-linear patterns.

\subsection{Deep learning models}

Deep learning models have advanced TS forecasting, starting with Recurrent Neural Networks (RNNs), specifically designed to model sequential data.
In particular, advanced variants such as RNNs with Long Short-Term Memory units, widely adopted within the TS community, have seen significantly increased usage \citep{hochreiter1997long}.
Additionally, MLP-based models, such as DLinear \citep{zeng_are_2022}, N-BEATS \citep{Oreshkin2020}, and N-Hits \citep{challu2023nhits} use MLP to learn the coefficients that produce both backcast and forecast outputs from their structure.

Originally from Natural Language Processing (NLP), the Transformer architecture is increasingly adapted for time series forecasting, often with modified attention layers to capture temporal dependencies, as seen in \Cref{sec:all_champs} and other prior works, which we describe in the following.
Informer \citep{zhou2021informer} and Pyaformer \citep{liu2022pyraformer} are transformer-based models that modify the attention mechanism. 
Informer designs a ProbSparse self-attention mechanism to replace the standard self-attention. 
Pyaformer, on the other hand, presents a pyramidal attention module, where the inter-scale tree structure captures features at different resolutions, and the intra-scale neighboring connections model the temporal dependencies across different ranges.
Wu et al. \citep{wu2021autoformer} introduced the Autoformer with an Auto-Correlation mechanism to capture the series-wise temporal dependencies based on the learned periods.
Following, FEDformer \citep{zhou2022fedformer} utilizes a mixture-of-expert framework to improve seasonal-trend decomposition and integrates Fourier and Wavelet-enhanced blocks to capture key structures in the TS.
\citep{zhang2023crossformer} presented Crossformer, a transformer-based model utilizing cross-dimension dependency for multivariate TS forecasting.
Another recent approach is TimesNet \citep{wu_timesnet_2023}, which is a univariate 2D CNN that segments 1D time series according to Fourier decomposition. 
The segments are then stacked to build a 2D series. 
This enables the convolutions to simultaneously look at the local structure of the signal at $t_i$ and $t_{i-T}$ simultaneously, where $T$ denotes a dominant signal period.

\subsection{Foundation Models}
\label{sec:foundational_models}
There is a growing interest in foundation models designed explicitly for TS tasks \citep{ShiMML24, yao2025towards}. 
Tiny Time Mixers \citep{ekambaram2024ttms} introduce a compact model for multivariate TS forecasting.
Timer-XL is a foundation model for unified time series forecasting, supporting univariate and multivariate data by extending next-token prediction for causal generation \citep{liu2024timer}.
The model introduced a universal TimeAttention mechanism to capture fine-grained intra- and inter-series dependencies.
MOIRAI \citep{woo2024unified} addresses challenges like cross-frequency learning and varied distributional properties in large-scale data, achieving competitive zero-shot forecasting performance.
TimeGPT-1 \citep{garza2023timegpt} and Lag-LLama \citep{rasul2023lag}, utilizing decoder-only transformer architectures and achieving strong zero-shot generalization.
Chronos \citep{ansari2024chronos} trains transformer-based models on discrete tokens processed from TS data, demonstrating superior performance on diverse datasets.

\subsection{Large Language Models}
\label{sec:llms}

The success of Large Language Models (LLMs) like BERT and GPT in NLP has inspired researchers to apply these models to TS tasks.
As outlined in \citep{jin_position_2024}, LLMs may serve in three roles (R): as Enhancers (R1), which incorporate domain-specific external knowledge while relying on specialized models for prediction; Forecasters (R2), which replace expert models entirely and cast LLMs directly as predictive models; or Agents (R3), which orchestrate workflows involving external tools and models. 
One significant approach involves transforming numerical TS data into natural language prompts to leverage pre-trained language models without modifications. 
PromptCast \citep{xue2023promptcast} and \citep{gruver2024large} present this method, demonstrating generalization in zero-shot settings and often outperforming traditional numerical models.
However, recent work cast doubts on the actual significance of LLMs as base forecasters \citep{tan_are_2024}.
Moving to few-shot training strategies, TEST \citep{sun2023test} adapts TS data for pre-trained LLMs by tokenizing the data and aligning the embedding space, particularly in few-shot and generalization scenarios.
Several frameworks focus on enhancing TS forecasting through specialized fine-tuning strategies such as LLM4TS \citep{chang2023llm4ts} and TEMPO \citep{cao2023tempo}.

\section{Datasets}
\label{sec:dataset_statistics}

We include a popular set of datasets (\textit{ETT*}, \textit{Electricity}, \textit{Weather}, \textit{Exchange}) and a set of larger datasets (\textit{MotorImagery}, \textit{TDBrain}, \textit{BeijingAir}, \textit{BenzeneConcentration}, \textit{AustraliaRainfall}, \textit{KDDCup2018}, \textit{PedestrianCounts}) representing a subset of the Unified Time Series Dataset (UTSD) \citep{liu2024timer}.

This section provides a summary of descriptive statistics about the employed datasets, an example of two datasets with clear and unclear patterns, respectively, and dataset-specific preprocessing steps.

\subsection{Dataset statistics}
\label{sec:dataset_stats}
\input{table_dataset_statistics}

In \Cref{tab:dataset_statistics}, we provide a comprehensive description of the datasets employed in this work and their corresponding statistics.
Next, we describe in detail the methodology to derive such dataset statistics.

\textbf{Time steps and channels.} We counted the \textit{number of time steps} and \textit{number of channels}.

\textbf{Shannon entropy and spectral entropy.} Shannon entropy quantifies the average level of uncertainty or information content associated with the outcomes in a discrete variable $X$.
Spectral entropy, a related concept, applies this measure to the frequency domain, using the normalized power spectral density as the probability distribution.
Entropy is calculated as 
\[
H(X) = -\sum_{x\in \chi} p(x) \log p(x)
\]
where $\chi$ is the set of all possible outcomes, and $p(x)$ is the probability of outcome $x$. 

\textbf{Sample entropy.} Sample entropy is a statistical measure used to quantify the complexity or regularity of a time series. 
Unlike Shannon entropy, which evaluates uncertainty in discrete probability distributions, sample entropy assesses the likelihood that similar patterns in the time series remain similar at the next time step. 
It is defined as the negative natural logarithm of the conditional probability that two sequences of length $m$ that match within a tolerance $r$ will still match when extended to $m+1$. 
A lower sample entropy indicates more regularity or predictability in the series, while a higher value suggests greater randomness or complexity.
Sample entropy is calculated as
\[
\text{SampEn}(m, r, N) = -\ln \frac{A}{B}
\]
where $m$ is the embedding dimension, $r$ is the tolerance, $N$ is the total length of the time series, $A$ is the number of matching pairs of length $m+1$, and $B$ is the number of matching pairs of length $m$.

\textbf{Stationarity.} In LTSF, stationarity is an important characteristic of time series, where the statistical characteristics, such as mean and variance, remain constant over time. 
We used the Augmented Dickey-Fuller (ADF) test to assess stationarity. 
This test evaluates the null hypothesis that a unit root is present and confirms stationarity if $\gamma < 0$ and the result is statistically significant.
We report $\gamma$ since it scales with stationarity.

\textbf{Complexity.} In time series analysis, complexity refers to the irregularity or unpredictability in the data.
We quantify complexity by using Higuchi's fractal dimension. 
A higher fractal dimension indicates greater complexity, while a lower value suggests more regularity or predictability in the data.
Higuchi’s fractal dimension is calculated as  
\[
D = \lim_{k \to 0} \frac{\log \left( \sum_{i=1}^n \left( \frac{|x_{i+k} - x_i|}{k} \right) \right)}{\log k}
\]
where $x_i$ is a point, $k$ is the time scale, and the sum is taken over different segments of the time series.

\textbf{Inter-variate similarity.} As a proxy for inter-variate similarity, we provide the explained variance of the first principal component (PC1) obtained through principal component analysis (PCA). 
PC1 represents the direction of maximum variance in the data, capturing the dominant shared variation among the variables. 
The explained variance of PC1 quantifies the proportion of the total variance that is accounted for by this component. 
A higher explained variance indicates stronger similarity and shared dynamics among the variables, while a lower value suggests more independent behavior.

\subsection{Clear vs. unclear patterns}
\label{sec:clear_vs_unclear_ds}

\begin{figure*}[t]
    \centering
    \includegraphics[scale=0.75]{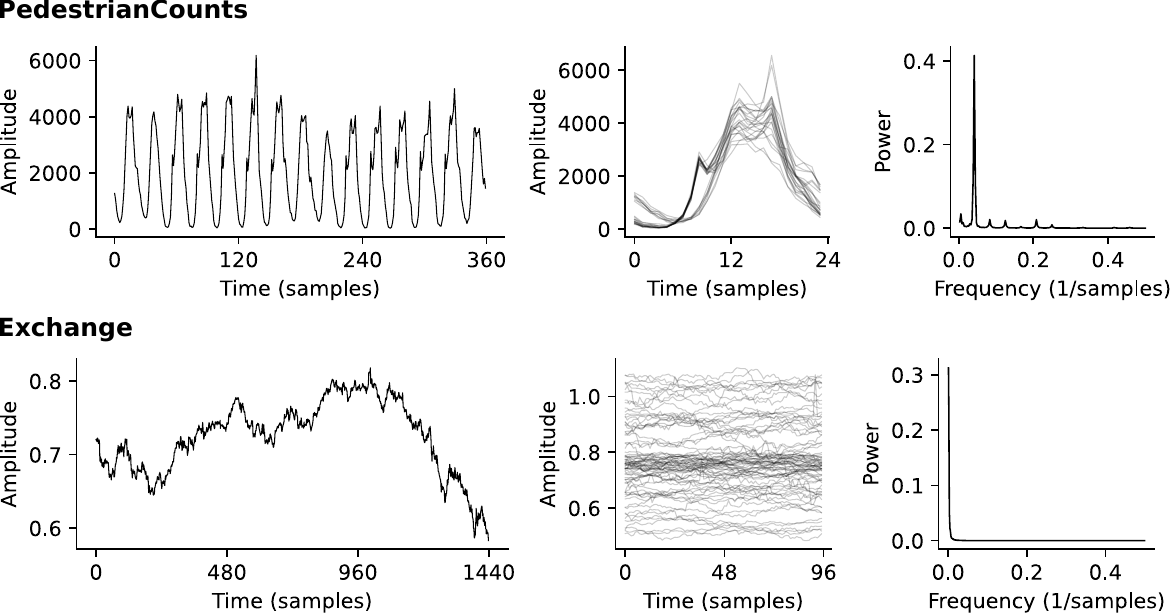}
    \caption{\textbf{Clear vs. unclear patterns (top vs bottom) in two datasets.}
    }
    \label{fig:data_patterns}
\end{figure*}

In \Cref{sec:all_similar}, we assessed the performance of a linear model versus a transformer model on datasets with clear and unclear patterns, respectively.
We use the same dataset as BasicTS+ \citep{shao2024exploring} with an unclear pattern (\textit{Exchange}) and replace their previously used PEMS with a clear pattern by \textit{PedestrianCounts} (\Cref{fig:data_patterns}).
The plots of the time series highlight contrasting characteristics: \textit{Exchange} displays seemingly random trends, whereas \textit{PedestrianCounts} exhibits evident cyclic behavior.
To further emphasize this distinction, we provide butterfly plots, which reveal pronounced periodic patterns in \textit{PedestrianCounts} and irregular, stochastic-like trends in \textit{Exchange}.
Additionally, the power spectrum analysis underscores this contrast, showing a dominant peak for \textit{PedestrianCounts} and an absence of such peaks for \textit{Exchange}.

\subsection{Preprocessing}

We followed the preprocessing steps from TSlib \citep{wang_deep_2024}. 
Furthermore, we added functionality to import data from UTSD as curated by \citep{liu2024timer}. 
Since the UTSD datasets are magnitudes larger, we modified the respective dataloader to return sequences at a stride length $S=100$ to accelerate the training.

\section{HP search details}
\label{sec:hp_search_appendix}

\textbf{HP search.} To ensure a fair comparison, we performed an extensive HP search for all models on all the datasets.
Specifically, we searched for an input length between 96 and 512, model size \(d_{m}\) from 16 to 512, learning rate ranging from \(10^{-5}\) to \(0.1\), and encoder layers between 1 and 5. The specific ranges are presented in \Cref{tab:exact_hyperparameters}.
TimeMixer was limited to a maximum of 3 layers and a model size of 128 due to increasingly high memory demands ($>49$GB  of VRAM on an RTX A6000 GPU if $d_{m} > 128$ and $L = 720$). 
However, this is unlikely to affect performance, as the original HP search for all datasets in this study yielded results within these limits.
Additionally, we limited the search space for $d_m$ in ModernTCN to a region closely centered around its default value for LTSF ($d_m=64$).
We used the Optuna framework \citep{optuna_2019} with a budget of 40 trials to optimize the HPs.
We employed the default \texttt{TPEsampler} for HP sampling and applied the \texttt{SuccessiveHalvingPruner}, configured with a minimum of three epochs and a reduction factor of two to prune unpromising trials.  
The search was conducted with a batch size of 8, a maximum of 15 epochs, and early stopping with a patience of 3 epochs.
All models were optimized with the Adam optimizer and an exponentially decaying scheduler following the default TSLib configuration. 
The optimal HPs, determined by the minimum validation loss in the trials, were used to train and evaluate the final model across three random seeds to ensure robust results.
We set the dimensions of the fully connected layers \(d_{f}\) equal to \(d_{m}\).
The patch length, a parameter used in PatchTST and TimeXer (and by extension also iPatch), was set based on the characteristics of the datasets (\Cref{tab:patch_length}).
As visualized in \Cref{fig:data_patterns}, there are datasets with dominant periodic behavior.
We set the patch length $P \approx \text{argmax}(\text{FFT}(X))$ wherever we observed such a natural pattern.
Unsurprisingly, the patch length resulted in a span of one day in all datasets with a pattern. 
In the remaining cases, we set the patch length manually (\textit{Exchange}, \textit{MotorImagery}, \textit{TDBrain}).
We aligned with the idea of TimesNet, which introduced series segmentation based on dominant frequencies to enhance performance \citep{wu_timesnet_2023}.
Moreover, the original works concluded that variations in patch length have minimal effects \citep{nie_time_2023, wang_timexer_2024}.
For the rest of the model HPs, we default to the configurations provided in TSLib.

\input{table_exact_hyperparameters}

\input{table_patch_lengths}

\section{Implementation reliability}
\label{sec:reliability_appendix}

To assess the reliability of our implementation, we compare our results (best and average MSE) against the original works on popular datasets. 
Despite not being directly comparable for slight differences in setups, we align with previous values (\Cref{tab:setup_reliability}).

\input{table_benchmark_setup_reliability_rebuttal}

\section{Stability of HP search}
\label{sec:hp_variability}

\begin{figure*}[t]
    \centering
    \begin{minipage}{0.48\textwidth}
        \centering
        \resizebox{\linewidth}{!}{\input{plots/hp_stability.pgf}}
    \end{minipage}
    \begin{minipage}{0.48\textwidth}
        \centering
        \resizebox{\linewidth}{!}{\input{plots/hp_stability_best.pgf}}
    \end{minipage}
    \caption{\textbf{HP search variability.}
    Comparison among two independent HP search results in terms of MSE for forecast horizon 96.
    The average of the final three models shows minimal variability except for a few cases (left), while the best model is even more stable (right).
    }
    \label{fig:hp_stability}
\end{figure*}

In this section, we perform a small proof-of-concept experiment to show the reliability of our HP search. 
For a subset of datasets and models, we performed two independent HP runs and consequent training of three models with the found optimal HPs to analyze the stability of the performed HP search.
We focused on forecast horizon 96.
In \Cref{fig:hp_stability}, we plot the MSE of the two searches over two opposing axes, meaning that points on the diagonal indicate a very stable search that leads to the same final result.
To provide an even more comprehensive analysis, we show the average of the three final models (left) and also the best model out of the final three (right).
The averages show a few cases with slight variability, namely PatchTST on \textit{BejingAir} and TimeXer on \textit{BenzeneConcentration}, but overall, the experiment proves a reliable and stable HP search across independent runs.
In the case of best results (\Cref{fig:hp_stability}, right), the variability is even less noticeable.

\begin{figure*}[t]
    \centering
    \resizebox{\linewidth}{!}
        {\input{plots/all_ds_violin.pgf}}
    \caption{\textbf{Violin plot for datasets and horizons.}
    We provide an alternative view of model performance on datasets and prediction horizons.
    Each point represents the average MSE obtained for a given forecast horizon over three independent seeds with the found HP configuration.
    It is clearly visible that the difference in average scores solely depends on the \textit{MotorImagery} dataset since baseline scores are comparable over the rest of the dataset-horizon setups.
    }
    \label{fig:all_ds_violin}
\end{figure*}

\section{Statistical tests}
\label{sec:stat_tests}

\subsection{Friedman test}

The Friedman test \citep{friedman1937use, friedman1940comparison} is a non-parametric statistical method designed as an alternative to repeated-measures ANOVA.
It enables the comparison of multiple algorithms across multiple datasets when the assumptions of parametric tests may not hold.
The test works by ranking the algorithms on each dataset separately, with the best-performing algorithm assigned a rank of 1, the second-best a rank of 2, and so forth.
In cases of ties, average ranks are assigned across the tied algorithms.

Let \( r^j_i \) denote the rank of the \( j \)-th algorithm out of \( k \) algorithms on the \( i \)-th dataset out of \( N \) datasets.
The Friedman test evaluates the average ranks of the algorithms, calculated as \(R^j = \frac{1}{N} \sum_{i=1}^N r^j_i\).
Under the null hypothesis, which assumes that all algorithms are equivalent in performance and thus their ranks \( R^j \) should be approximately equal, the Friedman statistic is given by:

\[
\chi^2_F = \frac{12N}{k(k+1)} \left[ \sum_{j=1}^k (R^j)^2 - \frac{k(k+1)^2}{4} \right]
\]

For sufficiently large values of \( N \) and \( k \), as a rule of thumb \( N > 10 \) and  \( k > 5 \) \citep{demvsar2006statistical}, this statistic follows a \( \chi^2 \) distribution with \( k-1 \) degrees of freedom.
Note our experimental setup aligns with these conditions.

The Friedman test, though less powerful than parametric repeated-measures ANOVA when its assumptions are met, is more robust in handling violations of these assumptions, with \citep{friedman1940comparison} observing largely consistent results between the two tests across 56 independent problems.
 
\subsection{Sign test}

The sign test \citep{salzberg1997comparing} is a non-parametric statistical method commonly used to compare the performance of two algorithms across multiple datasets.
It operates by evaluating the number of datasets on which each algorithm outperforms the other, assuming that the outcomes are independent and identically distributed.
Contrary to popular belief, counting only significant wins and losses actually makes the tests less reliable, as it imposes an arbitrary threshold of \( p < 0.05 \) to determine meaningfulness \citep{demvsar2006statistical}.

Under the null hypothesis, it is assumed that the two algorithms are equivalent in their performance, and thus, each algorithm has an equal probability (0.5) of outperforming the other on a given dataset. This leads to the number of wins for either algorithm following a binomial distribution with parameters \( N \) (the total number of datasets) and \( p = 0.5 \).
The null hypothesis is rejected if the observed number of wins for one algorithm is significantly different from \(\frac{N}{2}\), indicating that one algorithm systematically outperforms the other.

For small sample sizes, as in our case, with a total number of datasets being equal to 14,  critical values can be determined directly from the cumulative distribution function of the binomial distribution.
For larger sample sizes, the central limit theorem allows for an approximation using the normal distribution.
Specifically, the number of wins under the null hypothesis can be approximated by a normal distribution with mean \( \mu = N/2 \) and standard deviation \( \sigma = \sqrt{N}/2 \).
In cases where there are ties in performance, these ties are treated as supporting evidence for the null hypothesis.
To account for this, ties are split evenly between the two algorithms.
If the number of tied datasets is odd, one tie is disregarded to ensure that only whole numbers are assigned to each algorithm.

\section{Baseline comparisons}
\label{sec:baselines}

We provide an overview of included baselines for the top-performing LTSF models (\Cref{tab:baselines}).
We observe that previous models were replaced by recent ones as the field progressed.
We highlight that TimeMixer was not included as a baseline in TimeXer, although it was available at the time.

\input{table_baselines}

\section{Full results}
\label{sec:full_results}

\Cref{tab:full_results_mean} and \Cref{tab:full_results_min} shows the full results from our extensive experiments. 
We present the MSE and MAE for all forecast horizons $T \in \{96, 192, 336, 720\}$ and their average, respectively.
To provide a comprehensive view, we show the average (\Cref{tab:full_results_mean}) and the best (\Cref{tab:full_results_min}) values over three random seeds.
\Cref{tab:ipatch_results} presents the corresponding full results for iPatch.

For completeness, we also included two simple baselines to contextualize model performance in the challenging task of long-term forecasting. 
We included ARIMA as a classic statistical forecasting model \citep{BoxPierce1970} and Last Observation Carried Forward (LOCF), which predict all future time steps by repeating the last observed value from the input sequence \citep{hewamalage_forecast_2023}. 
We observe that, on average, the more recent machine learning models perform substantially better than the simple baselines (\Cref{tab:full_results_mean}, \Cref{tab:full_results_min}). 
However, ARIMA and LOCF are among the best-performing models on \textit{Exchange}. This is not surprising, given that stock market data lacks obvious periodicities and was previously shown to be best predicted by simple baselines \citep{hewamalage_forecast_2023}. 
This further supports our broader message: no model is consistently best, and performance can vary widely depending on the dataset.

\begin{table*}[t]
\caption{\textbf{Efficiency metrics.} 
Average efficiency metrics for all optimized models over datasets and prediction horizons (1,000 iterations, batch size of 1).
\textbf{\textcolor{red}{Best}} and {\textcolor{blue}{\underline{second-best}}} are highlighted.
}
\renewcommand{\arraystretch}{0.75}
\centering
\resizebox{\textwidth}{!}{
\begin{tabular}{lcccccccc}
\toprule
Metric & DLinear & PatchTST & iTransformer & TimeMixer & TimeXer & S-Mamba & xLSTMTime & ModernTCN \\
\midrule
\texttt{Train TP} ($\frac{\text{seq}}{\text{s}}$) 
& \textbf{\textcolor{red}{1663.58}} 
& 281.78 
& \textcolor{blue}{\underline{324.49}} 
& 92.82 
& 162.29 
& 68.57 
& 187.70 
& 97.30 \\

\texttt{Train memory} (MB) 
& \textbf{\textcolor{red}{150.86}} 
& 477.50 
& \textcolor{blue}{\underline{170.13}} 
& 380.41 
& 243.88 
& 456.86 
& 179.42 
& 1086.83 \\

\texttt{Test TP} ($\frac{\text{seq}}{\text{s}}$)
& \textbf{\textcolor{red}{4483.45}} 
& 879.69 
& \textcolor{blue}{\underline{1038.87}} 
& 311.98 
& 529.83 
& 246.29 
& 577.69 
& 242.11 \\

\texttt{Test memory} (MB)
& \textbf{\textcolor{red}{143.66}} 
& 440.95 
& \textcolor{blue}{\underline{155.21}} 
& 197.89 
& 176.21 
& 158.36 
& 167.00 
& 717.08 \\

\texttt{FLOPs} (G)
& \textbf{\textcolor{red}{0.02}} 
& 0.69 
& 0.20 
& 15.16 
& 4.36 
& 0.13 
& \textcolor{blue}{\underline{0.11}} 
& 4.21 \\

\texttt{\#params} (M)
& \textbf{\textcolor{red}{0.37}} 
& 2.37 
& \textcolor{blue}{\underline{1.54}} 
& 2.47 
& 3.76 
& 2.40 
& 2.15 
& 74.33 \\
\bottomrule
\end{tabular}

}
\label{tab:efficiency_metrics_avg}
\end{table*}

\input{table_full_results_mean}

\input{table_full_results_min}

\input{table_full_results_ipatch}

\section{Efficiency comparisons}
\label{sec:eff_appendix}

\textbf{Efficiency metrics setup.} The efficiency metrics were computed by evaluating model performance across 1,000 iterations using synthetic input and target data shaped according to the specified sequence length and prediction horizon stemming from the optimal experimental setup found during the HP search.
During each iteration, the model was executed in either training or inference mode with a batch size of one, depending on the configuration, and both the throughput (\texttt{TP}) and peak GPU memory (\texttt{memory}) usage were recorded.
The \texttt{TP} was quantified as the number of processed sequences per second, calculated by dividing the total number of sequences by the elapsed time.
We run the above analyses on a machine equipped with an RTX 4090 NVIDIA GPU and an AMD EPYC 7742 64-Core Processor (128 threads) CPU.
Additionally, we computed the number of floating-point operations (\texttt{FLOPs}) and total trainable parameters (\texttt{\#params}), offering insight into the theoretical computational complexity of evaluated models.
We subsequently scaled all efficiency metrics for interpretability: parameters to millions, FLOPs to gigaflops, and memory usage to megabytes (MB).

\textbf{Efficiency metrics results.} In \Cref{tab:efficiency_metrics_avg}, we provide an overview of the average efficiency metrics for all models over datasets and prediction horizons.
Note that for each dataset-horizon combination, we compute the metrics for the configuration found with the HP search.
Unsurprisingly, DLinear consistently demonstrates superior efficiency across all metrics, achieving the highest training and inference throughput, lowest memory usage, minimal FLOPs, and the smallest parameter count.
iTransformer is the second most efficient model in all metrics, except for \texttt{FLOPs} (xLSTMTime), showing an advantageous trade-off between throughput and memory across the more complex models.
TimeMixer and TimeXer lag particularly behind in terms of efficiency against DLinear.
Specifically, DLinear attains a training throughput of $\approx$1,600 sequences/s, making it roughly $10\times$ faster than TimeXer and $20\times$ faster than TimeMixer.
In terms of training memory, DLinear requires on average 150.86 MB, which is about $2.5\times$ less than TimeMixer and $1.6\times$ less than TimeXer. 

\end{document}

%% file: math_commands.tex
\usepackage{amsmath,amsfonts,bm}

\def\eqref#1{equation~\ref{#1}}

\def\1{\bm{1}}

\DeclareMathAlphabet{\mathsfit}{\encodingdefault}{\sfdefault}{m}{sl}
\SetMathAlphabet{\mathsfit}{bold}{\encodingdefault}{\sfdefault}{bx}{n}



%% file: table_main_results.tex
\begin{table*}[t]
\renewcommand{\arraystretch}{1.4}
\caption{\textbf{Results.}
Mean values averaged over prediction lengths.
Datasets span the energy, economy, transport, health, and environment domains. 
\textbf{\textcolor{red}{Best}} and {\textcolor{blue}{\underline{second-best}}} are highlighted.
}
\centering
\resizebox{\textwidth}{!}{

\begin{tabular}{c|cc|cc|cc|cc|cc|cc|cc|cc}
\hline
Model                & \multicolumn{2}{c|}{DLinear}                                                  & \multicolumn{2}{c|}{PatchTST}                                                 & \multicolumn{2}{c|}{iTransformer}                                       & \multicolumn{2}{c|}{TimeMixer}                                                & \multicolumn{2}{c|}{TimeXer}                                                  & \multicolumn{2}{c|}{S-Mamba}                                                   & \multicolumn{2}{c|}{xLSTMTime}                                                & \multicolumn{2}{c}{ModernTCN} \\ \hline
Metric               & MSE                                   & MAE                                   & MSE                                   & MAE                                   & MSE                                & MAE                                & MSE                                   & MAE                                   & MSE                                   & MAE                                   & MSE                                   & MAE                                   & MSE                                   & MAE                                   & MSE           & MAE           \\ \hline
ETTh1                & 0.474                                 & 0.477                                 & {\color[HTML]{FF0000} \textbf{0.414}} & {\color[HTML]{FF0000} \textbf{0.432}} & {\color[HTML]{0000FF} {\ul 0.424}} & 0.443                              & 0.429                                 & 0.443                                 & 0.425                                 & {\color[HTML]{0000FF} {\ul 0.44}}     & 0.467                                 & 0.466                                 & 0.53                                  & 0.515                                 & 0.532         & 0.51          \\
ETTm1                & {\color[HTML]{0000FF} {\ul 0.403}}    & {\color[HTML]{FF0000} \textbf{0.422}} & {\color[HTML]{FF0000} \textbf{0.394}} & {\color[HTML]{0000FF} {\ul 0.431}}    & 0.408                              & 0.437                              & 0.432                                 & 0.449                                 & 0.412                                 & 0.445                                 & 0.409                                 & 0.434                                 & 0.434                                 & 0.449                                 & 0.414         & 0.435         \\
ETTh2                & 0.485                                 & 0.474                                 & 0.379                                 & 0.41                                  & 0.378                              & {\color[HTML]{0000FF} {\ul 0.405}} & {\color[HTML]{FF0000} \textbf{0.374}} & {\color[HTML]{FF0000} \textbf{0.404}} & {\color[HTML]{0000FF} {\ul 0.377}}    & 0.406                                 & 0.383                                 & 0.407                                 & 0.803                                 & 0.647                                 & 0.405         & 0.423         \\
ETTm2                & {\color[HTML]{FF0000} \textbf{0.151}} & {\color[HTML]{FF0000} \textbf{0.261}} & {\color[HTML]{0000FF} {\ul 0.156}}    & {\color[HTML]{0000FF} {\ul 0.269}}    & 0.166                              & 0.276                              & 0.162                                 & 0.27                                  & 0.168                                 & 0.277                                 & 0.168                                 & 0.279                                 & 0.17                                  & 0.278                                 & 0.166         & 0.276         \\
Electricity          & {\color[HTML]{0000FF} {\ul 0.162}}    & 0.259                                 & 0.164                                 & 0.261                                 & 0.165                              & 0.262                              & {\color[HTML]{FF0000} \textbf{0.156}} & {\color[HTML]{FF0000} \textbf{0.254}} & 0.17                                  & 0.268                                 & 0.166                                 & 0.263                                 & 0.163                                 & {\color[HTML]{0000FF} {\ul 0.257}}    & 0.275         & 0.376         \\
Weather              & 0.244                                 & 0.298                                 & {\color[HTML]{0000FF} {\ul 0.225}}    & {\color[HTML]{FF0000} \textbf{0.263}} & 0.238                              & 0.279                              & 0.231                                 & 0.271                                 & {\color[HTML]{FF0000} \textbf{0.224}} & {\color[HTML]{0000FF} {\ul 0.264}}    & 0.237                                 & 0.277                                 & 0.229                                 & 0.268                                 & 0.241         & 0.282         \\
Exchange             & 0.389                                 & 0.434                                 & {\color[HTML]{FF0000} \textbf{0.368}} & {\color[HTML]{FF0000} \textbf{0.409}} & 0.379                              & 0.416                              & 0.55                                  & 0.476                                 & {\color[HTML]{0000FF} {\ul 0.374}}    & {\color[HTML]{0000FF} {\ul 0.412}}    & 0.426                                 & 0.438                                 & 0.699                                 & 0.6                                   & 0.761         & 0.531         \\
MotorImagery         & 4.592                                 & 1.183                                 & 3.775                                 & 1.006                                 & {\color[HTML]{0000FF} {\ul 1.692}} & {\color[HTML]{0000FF} {\ul 0.383}} & 3.869                                 & 1.011                                 & 3.649                                 & 0.915                                 & {\color[HTML]{FF0000} \textbf{0.745}} & {\color[HTML]{FF0000} \textbf{0.244}} & 6.622                                 & 1.362                                 & 3.088         & 0.769         \\
TDBrain              & 1.151                                 & 0.802                                 & 0.982                                 & 0.725                                 & 0.978                              & 0.722                              & 0.981                                 & 0.723                                 & {\color[HTML]{FF0000} \textbf{0.97}}  & {\color[HTML]{FF0000} \textbf{0.719}} & {\color[HTML]{0000FF} {\ul 0.972}}    & {\color[HTML]{0000FF} {\ul 0.72}}     & 1.211                                 & 0.799                                 & 0.984         & 0.725         \\
BeijingAir           & 0.583                                 & 0.472                                 & 0.578                                 & 0.46                                  & 0.58                               & 0.464                              & 0.592                                 & 0.471                                 & 0.582                                 & 0.462                                 & {\color[HTML]{FF0000} \textbf{0.567}} & {\color[HTML]{0000FF} {\ul 0.452}}    & {\color[HTML]{0000FF} {\ul 0.569}}    & {\color[HTML]{FF0000} \textbf{0.43}}  & 0.581         & 0.461         \\
BenzeneConcentration & {\color[HTML]{FF0000} \textbf{0.008}} & {\color[HTML]{FF0000} \textbf{0.02}}  & 0.011                                 & 0.042                                 & 0.012                              & 0.053                              & {\color[HTML]{FF0000} \textbf{0.008}} & {\color[HTML]{0000FF} {\ul 0.022}}    & 0.012                                 & 0.037                                 & {\color[HTML]{0000FF} {\ul 0.01}}     & 0.044                                 & {\color[HTML]{0000FF} {\ul 0.01}}     & 0.028                                 & 0.15          & 0.265         \\
AustraliaRainfall    & {\color[HTML]{0000FF} {\ul 0.838}}    & {\color[HTML]{0000FF} {\ul 0.751}}    & 0.855                                 & 0.757                                 & 0.849                              & 0.754                              & 0.853                                 & 0.755                                 & 0.847                                 & 0.753                                 & 0.848                                 & 0.753                                 & {\color[HTML]{FF0000} \textbf{0.827}} & {\color[HTML]{FF0000} \textbf{0.743}} & 0.872         & 0.766         \\
KDDCup2018           & {\color[HTML]{FF0000} \textbf{0.997}} & {\color[HTML]{0000FF} {\ul 0.63}}     & 1.086                                 & 0.647                                 & 1.088                              & 0.648                              & {\color[HTML]{0000FF} {\ul 1.035}}    & 0.631                                 & 1.045                                 & 0.64                                  & 1.085                                 & 0.646                                 & 1.051                                 & {\color[HTML]{FF0000} \textbf{0.628}} & 1.068         & 0.638         \\
PedestrianCounts     & 0.298                                 & 0.289                                 & {\color[HTML]{0000FF} {\ul 0.291}}    & 0.293                                 & 0.295                              & 0.285                              & 0.292                                 & 0.284                                 & 0.311                                 & 0.307                                 & 0.292                                 & {\color[HTML]{0000FF} {\ul 0.278}}    & {\color[HTML]{FF0000} \textbf{0.282}} & {\color[HTML]{FF0000} \textbf{0.259}} & 0.466         & 0.408         \\ \hline
Average              & 0.77                                  & 0.484                                 & 0.691                                 & 0.458                                 & {\color[HTML]{0000FF} {\ul 0.547}} & {\color[HTML]{0000FF} {\ul 0.416}} & 0.712                                 & 0.462                                 & 0.683                                 & 0.453                                 & {\color[HTML]{FF0000} \textbf{0.484}} & {\color[HTML]{FF0000} \textbf{0.407}} & 0.971                                 & 0.519                                 & 0.714         & 0.49          \\ \hline
Rank                 & 4.43                                  & 4.5                                   & {\color[HTML]{FF0000} \textbf{3.5}}   & {\color[HTML]{FF0000} \textbf{3.93}}  & 4.29                               & 4.5                                & {\color[HTML]{0000FF} {\ul 4.07}}     & {\color[HTML]{0000FF} {\ul 4.14}}     & {\color[HTML]{0000FF} {\ul 4.07}}     & {\color[HTML]{0000FF} {\ul 4.14}}     & {\color[HTML]{0000FF} {\ul 4.07}}     & {\color[HTML]{0000FF} {\ul 4.14}}     & 5.07                                  & 4.64                                  & 6.5           & 6             \\ \hline
\end{tabular}

}

\label{tab:main_results}
\end{table*}

%% file: table_dataset_statistics.tex
\begin{table*}[t]
\renewcommand{\arraystretch}{1.5}
\centering
\caption{\textbf{Dataset statistics.} Refer to \Cref{sec:dataset_stats} for a detailed description of the statistics.}
\vspace{0.1cm}
\resizebox{\textwidth}{!}{
\begin{tabular}{ccccccccccccc}
\toprule
Domain      & Dataset              & \# Timesteps & \# Channels & Shannon Entr. & Spectral Entr. & Sample Entr. & Stationarity & Complexity & Expl. Var. & Source \\ 
\midrule
Energy      & ETTh1                & 17420        & 7           & 0.775           & 0.669            & 0.769          & -5.909       & 0.497      & 0.344                    &    \cite{wang_deep_2024}    \\
Energy      & ETTm1                & 69680        & 7           & 0.789           & 0.548            & 0.430           & -14.985      & 0.485      & 0.346                    &    \cite{wang_deep_2024}    \\
Energy      & ETTh2                & 17420        & 7           & 0.813           & 0.639            & 0.526          & -4.136       & 0.397      & 0.431                    &    \cite{wang_deep_2024}    \\
Energy      & ETTm2                & 69680        & 7           & 0.817           & 0.527            & 0.319          & -5.664       & 0.425      & 0.431                    &    \cite{wang_deep_2024}    \\
Energy      & Electricity          & 26304        & 321         & 0.516           & 0.497            & 0.714          & -8.445       & 0.673      & 0.547                    &    \cite{wang_deep_2024}    \\
Environment & Weather              & 52696        & 21          & 0.453           & 0.57             & 0.110           & -26.681      & 0.632      & 0.424                    &    \cite{wang_deep_2024}    \\
Economic    & Exchange             & 7588         & 8           & 0.805           & 0.347            & 0.066          & -1.902       & 0.529      & 0.618                    &    \cite{wang_deep_2024}    \\
Health      & MotorImagery         & 1134000      & 64          & 0.719           & 0.519            & 0.326          & -3.133       & 0.763      & 0.305                    &    \cite{liu2024timer}    \\
Health      & TDBrain              & 2221212      & 33          & 0.823           & 0.749            & 0.987          & -3.167       & 0.404      & 0.475                    &    \cite{liu2024timer}    \\
Environment & BeijingAir           & 407184       & 9           & 0.493           & 0.686            & 0.951          & -13.253      & 0.165      & 0.383                    &    \cite{liu2024timer}    \\
Environment & BenzeneConcentration & 2042880      & 8           & 0.799           & 0.701            & 1.938          & -3.114       & -0.049     & 0.534                    &    \cite{liu2024timer}    \\
Environment & AustraliaRainfall    & 3846408      & 3           & 0.838           & 0.604            & 2.215          & -31.734      & -0.013     & 0.996                    &    \cite{liu2024timer}    \\
Environment & KDDCup2018           & 2942364      & 1           & 0.569           & 0.665            & 0.410           & -9.379       & 0.530       & 1.000                        &    \cite{liu2024timer}    \\
Transport   & PedestrianCounts     & 3132346      & 1           & 0.687           & 0.522            & 0.412          & -4.590        & 0.630       & 1.000                        &    \cite{liu2024timer}    \\ \bottomrule
\end{tabular}
}
\label{tab:dataset_statistics}
\end{table*}

%% file: table_exact_hyperparameters.tex
\begin{table}[h]
\caption{\textbf{Searched HPs values.} }
\small
\centering
\begin{tabular}{@{}
>{\columncolor[HTML]{FFFFFF}}c 
>{\columncolor[HTML]{FFFFFF}}c 
>{\columncolor[HTML]{FFFFFF}}c 
>{\columncolor[HTML]{FFFFFF}}c @{}}
\toprule
Hyperparameter         & TimeMixer                & ModernTCN                & Other Models             \\ \midrule
Input Length           & \{96, 192, 336, 720\}                   & \{96, 192, 336, 720\}                   & \{96, 192, 336, 720\}                   \\
Learning Rate          & \{\(10^{-5}, 10^{-4}, 10^{-3}, 10^{-2}\)\} & \{\(10^{-5}, 10^{-4}, 10^{-3}, 10^{-2}\)\} & \{\(10^{-5}, 10^{-4}, 10^{-3}, 10^{-2}\)\} \\
\#Layers & \{1, 2, 3\}                             & \{1, 2, 3, 4\}                           & \{1, 2, 3, 4\}                           \\
\(d_{m}\)               & \{16, 32, 64, 128\}                     & \{32, 64, 128, 256\}                     & \{16, 32, 64, 128, 256, 512\}           \\ \bottomrule
\end{tabular}
\label{tab:exact_hyperparameters}
\end{table}

%% file: table_patch_lengths.tex
\begin{table*}[h]
\caption{\textbf{Patch lengths.} 
Employed patch lengths for PatchTST, TimeXer, and iPatch models. 
}
\renewcommand{\arraystretch}{1.8}
\centering
\resizebox{\textwidth}{!}{
\begin{tabular}{ccccccccccccc}
\toprule
Dataset      & \rotatebox{60}{ETTh*} & \rotatebox{60}{ETTm*} & \rotatebox{60}{Electricity} & \rotatebox{60}{Weather} & \rotatebox{60}{Exchange} & \rotatebox{60}{MotorImagery} & \rotatebox{60}{TDBrain} & \rotatebox{60}{BeijingAir} & \rotatebox{60}{BenzeneConc.} & \rotatebox{60}{AustraliaRainf.} & \rotatebox{60}{KDDCup2018} & \rotatebox{60}{PedestrianC.} \\ \midrule
Patch Length & 24    & 96    & 24          & 24      & 96       & 96           & 48      & 24         & 24                   & 24                & 24         & 24               \\ \bottomrule
\end{tabular}
}
\label{tab:patch_length}
\end{table*}

%% file: table_benchmark_setup_reliability_rebuttal.tex
\begin{table*}[h]
\caption{\textbf{Implementation reliability.}
To assess the reliability of our implementation, we compare our results (best and average MSE) against the original works on popular datasets.
Despite not being directly comparable due to slight differences in experimental setups, we align with previous values.
}
\centering
\small 
\begin{tabularx}{\textwidth}{X*4{>{\centering\arraybackslash}X}@{}}
\toprule
Model & Dataset & Original & Ours Min & Ours Mean \\ \midrule

\multirow{3}{*}{DLinear} 
& ETT*        & 0.370 & 0.378 & 0.378 \\
& Electricity & 0.166 & 0.162 & 0.162 \\
& Weather     & 0.246 & 0.244 & 0.244 \\ \midrule

\multirow{3}{*}{PatchTST} 
& ETT*        & 0.338 & 0.332 & 0.336 \\
& Electricity & 0.159 & 0.163 & 0.164 \\
& Weather     & 0.225 & 0.224 & 0.225 \\ \midrule

\multirow{3}{*}{iTransformer} 
& ETT*        & 0.383 & 0.342 & 0.344 \\
& Electricity & 0.178 & 0.162 & 0.166 \\
& Weather     & 0.258 & 0.237 & 0.239 \\ \midrule

\multirow{3}{*}{TimeMixer} 
& ETT*        & 0.333 & 0.342 & 0.349 \\
& Electricity & 0.156 & 0.154 & 0.156 \\
& Weather     & 0.222 & 0.225 & 0.231 \\ \midrule

\multirow{3}{*}{TimeXer} 
& ETT*        & 0.363 & 0.341 & 0.346 \\
& Electricity & 0.171 & 0.165 & 0.170 \\
& Weather     & 0.241 & 0.222 & 0.224 \\ \midrule

\multirow{3}{*}{ModernTCN} 
& ETT*        & 0.333 & 0.374 & 0.379 \\
& Electricity & 0.156 & 0.268 & 0.275 \\
& Weather     & 0.224 & 0.233 & 0.241 \\ \midrule

\multirow{3}{*}{xLSTMTime} 
& ETT*        & 0.339 & 0.460 & 0.484 \\
& Electricity & 0.157 & 0.160 & 0.163 \\
& Weather     & 0.221 & 0.226 & 0.229 \\ \midrule

\multirow{3}{*}{S-Mamba} 
& ETT*        & 0.380 & 0.353 & 0.357 \\
& Electricity & 0.170 & 0.164 & 0.166 \\
& Weather     & 0.251 & 0.230 & 0.237 \\

\bottomrule
\end{tabularx}
\label{tab:setup_reliability}
\end{table*}

%% file: table_baselines.tex
\begin{table*}[h]
    \centering
    \caption[caption]{\textbf{Included baseline models from top-performing models in long-term forecasting}. x:~included; \textbf{o}:~introduced}
        \small
    \begin{tabular}{l|ccccc|cccccccccccc}
        \toprule \noalign{\vskip 1mm}  
        Model                      & \rotatebox{90}{DLinear \cite{zeng_are_2022}}    & \rotatebox{90}{PatchTST \cite{nie_time_2023}}    & \rotatebox{90}{TimeMixer \cite{wang_timemixer_2024}}    & \rotatebox{90}{iTransformer \cite{liu_itransformer_2024}}    & \rotatebox{90}{TimeXer \cite{wang_timexer_2024}}    & \rotatebox{90}{FEDformer \cite{zhou2022fedformer}} & \rotatebox{90}{Autoformer \cite{wu2021autoformer}} & \rotatebox{90}{Informer \cite{zhou2021informer}} & \rotatebox{90}{Pyraformer \cite{liu2022pyraformer}} & \rotatebox{90}{LogTrans \cite{li_enhancing_2019}} & \rotatebox{90}{Stationary \cite{liu2022stationary}} & \rotatebox{90}{Crossformer \cite{zhang2023crossformer}} & \rotatebox{90}{TimesNet \cite{wu_timesnet_2023}}   & \rotatebox{90}{SCINet \cite{liu2022scinet}} & \rotatebox{90}{Rlinear \cite{li_rlinear_2023}} & \rotatebox{90}{TiDE \cite{das_long-term_2024}} & \rotatebox{90}{\textit{others}} \\ \midrule 
        DLinear \cite{zeng_are_2022}                      & \textbf{o} &            &            &            &            & x       & x       & x       & x        & x        &          &          &          &                              &                              &                                     \\
         PatchTST \cite{nie_time_2023}                       & x          & \textbf{o} &            &            &            & x       & x       & x       & x        & x        &          &          &          &                              &                              &                                     \\
        TimeMixer \cite{wang_timemixer_2024}                       & x          & x          & \textbf{o} &            &            & x       & x       & x       &          &          & x        & x        & x        &          &                              &                              & x                                   \\
        iTransformer \cite{liu_itransformer_2024}                       & x          & x          &            & \textbf{o} &            & x       & x       &         &          &          & x        & x        & x        & x        & x                            & x                            & \multicolumn{1}{l}{}                \\
    TimeXer \cite{wang_timexer_2024}                    & x          & x          &            & x          & \textbf{o} &         & x       &         & x        &          & x        & x        & x        & x        & x                            & x                            &    \\ \bottomrule     \end{tabular}
    \label{tab:baselines}
\end{table*}

%% file: table_full_results_mean.tex
\begin{table*}[t]
\tiny
\renewcommand{\arraystretch}{1.4}

\setlength{\tabcolsep}{2pt}
\caption{\textbf{Full results.}
\underline{Mean values} for all prediction lengths. \textbf{\textcolor{red}{Best}} and {\textcolor{blue}{\underline{second-best}}} are highlighted.
}
\centering
\resizebox{\textwidth}{!}{
%
}             & Avg    & 4.592                                 & 1.183                                 & 3.775                                 & 1.006                                 & {\color[HTML]{0000FF} {\ul 1.692}} & {\color[HTML]{0000FF} {\ul 0.383}}    & 3.869                                 & 1.011                                 & 3.649                                 & 0.915                                 & {\color[HTML]{FF0000} \textbf{0.745}} & {\color[HTML]{FF0000} \textbf{0.244}} & 6.622                                 & 1.362                                 & 3.088                                 & 0.769                                 & 5.818                              & 1.321                                 & 4.016                                 & 1.115                                 \\ \hline
                                                                                      & 96     & 0.769                                 & 0.65                                  & {\color[HTML]{0000FF} {\ul 0.664}}    & {\color[HTML]{0000FF} {\ul 0.596}}    & 0.673                              & 0.598                                 & 0.673                                 & {\color[HTML]{0000FF} {\ul 0.596}}    & {\color[HTML]{FF0000} \textbf{0.657}} & {\color[HTML]{FF0000} \textbf{0.591}} & 0.67                                  & {\color[HTML]{0000FF} {\ul 0.596}}    & 0.87                                  & 0.657                                 & 0.691                                 & 0.604                                 & 0.863                              & 0.675                                 & 1.106                                 & 0.764                                 \\
                                                                                      & 192    & 1.002                                 & 0.746                                 & 0.835                                 & 0.669                                 & 0.831                              & 0.667                                 & 0.837                                 & 0.669                                 & {\color[HTML]{FF0000} \textbf{0.824}} & {\color[HTML]{FF0000} \textbf{0.663}} & {\color[HTML]{0000FF} {\ul 0.827}}    & {\color[HTML]{0000FF} {\ul 0.664}}    & 1.135                                 & 0.759                                 & 0.843                                 & 0.672                                 & 1.013                              & 0.739                                 & 1.261                                 & 0.826                                 \\
                                                                                      & 336    & 1.286                                 & 0.846                                 & 1.066                                 & 0.754                                 & 1.053                              & 0.749                                 & 1.057                                 & 0.75                                  & {\color[HTML]{0000FF} {\ul 1.05}}     & {\color[HTML]{0000FF} {\ul 0.748}}    & {\color[HTML]{FF0000} \textbf{1.042}} & {\color[HTML]{FF0000} \textbf{0.744}} & 1.317                                 & 0.836                                 & 1.056                                 & 0.751                                 & 1.207                              & 0.809                                 & 1.467                                 & 0.895                                 \\
                                                                                      & 720    & 1.547                                 & 0.964                                 & 1.363                                 & 0.88                                  & 1.356                              & 0.876                                 & 1.358                                 & 0.877                                 & {\color[HTML]{0000FF} {\ul 1.348}}    & {\color[HTML]{0000FF} {\ul 0.872}}    & 1.349                                 & 0.874                                 & 1.522                                 & 0.946                                 & {\color[HTML]{FF0000} \textbf{1.344}} & {\color[HTML]{FF0000} \textbf{0.871}} & 1.495                              & 0.92                                  & 1.782                                 & 1.007                                 \\ \cline{2-22} 
\multirow{-5}{*}{TDBrain}                                                             & Avg    & 1.151                                 & 0.802                                 & 0.982                                 & 0.725                                 & 0.978                              & 0.722                                 & 0.981                                 & 0.723                                 & {\color[HTML]{FF0000} \textbf{0.97}}  & {\color[HTML]{FF0000} \textbf{0.719}} & {\color[HTML]{0000FF} {\ul 0.972}}    & {\color[HTML]{0000FF} {\ul 0.72}}     & 1.211                                 & 0.799                                 & 0.984                                 & 0.725                                 & 1.145                              & 0.786                                 & 1.404                                 & 0.873                                 \\ \hline
                                                                                      & 96     & 0.529                                 & 0.441                                 & {\color[HTML]{FF0000} \textbf{0.522}} & {\color[HTML]{0000FF} {\ul 0.43}}     & 0.54                               & 0.441                                 & 0.551                                 & 0.452                                 & {\color[HTML]{0000FF} {\ul 0.526}}    & 0.432                                 & 0.552                                 & 0.442                                 & 0.537                                 & {\color[HTML]{FF0000} \textbf{0.41}}  & 0.527                                 & 0.435                                 & 0.665                              & 0.504                                 & 1.16                                  & 0.598                                 \\
                                                                                      & 192    & 0.569                                 & 0.463                                 & 0.572                                 & 0.454                                 & 0.571                              & 0.458                                 & 0.595                                 & 0.472                                 & 0.574                                 & 0.454                                 & {\color[HTML]{0000FF} {\ul 0.564}}    & {\color[HTML]{0000FF} {\ul 0.453}}    & {\color[HTML]{FF0000} \textbf{0.557}} & {\color[HTML]{FF0000} \textbf{0.422}} & 0.567                                 & 0.454                                 & 0.74                               & 0.541                                 & 1.214                                 & 0.646                                 \\
                                                                                      & 336    & 0.591                                 & 0.473                                 & 0.591                                 & 0.466                                 & 0.595                              & 0.468                                 & 0.589                                 & 0.467                                 & 0.588                                 & 0.465                                 & {\color[HTML]{0000FF} {\ul 0.584}}    & {\color[HTML]{0000FF} {\ul 0.461}}    & {\color[HTML]{FF0000} \textbf{0.572}} & {\color[HTML]{FF0000} \textbf{0.435}} & 0.588                                 & 0.463                                 & 0.756                              & 0.548                                 & 1.077                                 & 0.628                                 \\
                                                                                      & 720    & 0.643                                 & 0.509                                 & 0.627                                 & 0.488                                 & 0.615                              & 0.488                                 & 0.631                                 & 0.495                                 & 0.639                                 & 0.496                                 & {\color[HTML]{FF0000} \textbf{0.567}} & {\color[HTML]{0000FF} {\ul 0.453}}    & {\color[HTML]{0000FF} {\ul 0.608}}    & {\color[HTML]{FF0000} \textbf{0.452}} & 0.64                                  & 0.491                                 & 0.788                              & 0.571                                 & 1.149                                 & 0.653                                 \\ \cline{2-22} 
\multirow{-5}{*}{BeijingAir}                                                          & Avg    & 0.583                                 & 0.472                                 & 0.578                                 & 0.46                                  & 0.58                               & 0.464                                 & 0.592                                 & 0.471                                 & 0.582                                 & 0.462                                 & {\color[HTML]{FF0000} \textbf{0.567}} & {\color[HTML]{0000FF} {\ul 0.452}}    & {\color[HTML]{0000FF} {\ul 0.569}}    & {\color[HTML]{FF0000} \textbf{0.43}}  & 0.581                                 & 0.461                                 & 0.737                              & 0.541                                 & 1.15                                  & 0.631                                 \\ \hline
                                                                                      & 96     & {\color[HTML]{0000FF} {\ul 0.007}}    & {\color[HTML]{FF0000} \textbf{0.016}} & 0.01                                  & 0.041                                 & 0.01                               & 0.058                                 & {\color[HTML]{FF0000} \textbf{0.006}} & {\color[HTML]{FF0000} \textbf{0.016}} & 0.009                                 & 0.041                                 & {\color[HTML]{0000FF} {\ul 0.007}}    & {\color[HTML]{0000FF} {\ul 0.037}}    & 0.008                                 & {\color[HTML]{FF0000} \textbf{0.016}} & 0.281                                 & 0.39                                  & 0.931                              & 0.764                                 & 1.264                                 & 0.862                                 \\
                                                                                      & 192    & {\color[HTML]{0000FF} {\ul 0.007}}    & {\color[HTML]{FF0000} \textbf{0.015}} & 0.01                                  & 0.027                                 & 0.008                              & 0.037                                 & {\color[HTML]{FF0000} \textbf{0.006}} & {\color[HTML]{0000FF} {\ul 0.019}}    & 0.01                                  & 0.037                                 & 0.009                                 & 0.044                                 & 0.008                                 & 0.028                                 & 0.055                                 & 0.157                                 & 1.003                              & 0.81                                  & 1.345                                 & 0.9                                   \\
                                                                                      & 336    & {\color[HTML]{FF0000} \textbf{0.008}} & {\color[HTML]{FF0000} \textbf{0.019}} & {\color[HTML]{0000FF} {\ul 0.009}}    & 0.042                                 & 0.013                              & 0.059                                 & {\color[HTML]{FF0000} \textbf{0.008}} & 0.028                                 & 0.011                                 & 0.032                                 & 0.011                                 & 0.053                                 & {\color[HTML]{0000FF} {\ul 0.009}}    & {\color[HTML]{0000FF} {\ul 0.024}}    & 0.131                                 & 0.258                                 & 1.043                              & 0.829                                 & 1.281                                 & 0.87                                  \\
                                                                                      & 720    & {\color[HTML]{FF0000} \textbf{0.011}} & {\color[HTML]{0000FF} {\ul 0.028}}    & 0.016                                 & 0.058                                 & 0.015                              & 0.058                                 & {\color[HTML]{0000FF} {\ul 0.013}}    & {\color[HTML]{FF0000} \textbf{0.027}} & 0.016                                 & 0.038                                 & 0.014                                 & 0.043                                 & 0.015                                 & 0.041                                 & 0.132                                 & 0.256                                 & 1.104                              & 0.856                                 & 1.281                                 & 0.871                                 \\ \cline{2-22} 
\multirow{-5}{*}{\begin{tabular}[c]{@{}c@{}}Benzene\\ Concen-\\ tration\end{tabular}} & Avg    & {\color[HTML]{FF0000} \textbf{0.008}} & {\color[HTML]{FF0000} \textbf{0.02}}  & 0.011                                 & 0.042                                 & 0.012                              & 0.053                                 & {\color[HTML]{FF0000} \textbf{0.008}} & {\color[HTML]{0000FF} {\ul 0.022}}    & 0.012                                 & 0.037                                 & {\color[HTML]{0000FF} {\ul 0.01}}     & 0.044                                 & {\color[HTML]{0000FF} {\ul 0.01}}     & 0.028                                 & 0.15                                  & 0.265                                 & 1.02                               & 0.815                                 & 1.293                                 & 0.876                                 \\ \hline
                                                                                      & 96     & {\color[HTML]{0000FF} {\ul 0.806}}    & 0.73                                  & 0.815                                 & 0.732                                 & 0.809                              & 0.727                                 & 0.813                                 & 0.729                                 & 0.807                                 & 0.728                                 & 0.808                                 & {\color[HTML]{0000FF} {\ul 0.726}}    & {\color[HTML]{FF0000} \textbf{0.788}} & {\color[HTML]{FF0000} \textbf{0.716}} & 0.84                                  & 0.745                                 & 1.911                              & 1.092                                 & 1.878                                 & 1.083                                 \\
                                                                                      & 192    & {\color[HTML]{0000FF} {\ul 0.839}}    & {\color[HTML]{0000FF} {\ul 0.751}}    & 0.862                                 & 0.758                                 & 0.851                              & 0.755                                 & 0.855                                 & 0.755                                 & 0.847                                 & 0.753                                 & 0.847                                 & 0.752                                 & {\color[HTML]{FF0000} \textbf{0.826}} & {\color[HTML]{FF0000} \textbf{0.743}} & 0.867                                 & 0.762                                 & 1.978                              & 1.122                                 & 1.949                                 & 1.115                                 \\
                                                                                      & 336    & {\color[HTML]{0000FF} {\ul 0.85}}     & {\color[HTML]{0000FF} {\ul 0.76}}     & 0.864                                 & 0.764                                 & 0.863                              & 0.763                                 & 0.868                                 & 0.765                                 & 0.861                                 & 0.762                                 & 0.863                                 & 0.764                                 & {\color[HTML]{FF0000} \textbf{0.841}} & {\color[HTML]{FF0000} \textbf{0.754}} & 0.888                                 & 0.777                                 & 1.976                              & 1.128                                 & 1.944                                 & 1.119                                 \\
                                                                                      & 720    & {\color[HTML]{0000FF} {\ul 0.858}}    & {\color[HTML]{0000FF} {\ul 0.764}}    & 0.877                                 & 0.772                                 & 0.874                              & 0.77                                  & 0.876                                 & 0.771                                 & 0.871                                 & 0.769                                 & 0.873                                 & 0.769                                 & {\color[HTML]{FF0000} \textbf{0.853}} & {\color[HTML]{FF0000} \textbf{0.761}} & 0.893                                 & 0.78                                  & 2.011                              & 1.141                                 & 1.979                                 & 1.132                                 \\ \cline{2-22} 
\multirow{-5}{*}{\begin{tabular}[c]{@{}c@{}}Australia\\ Rainfall\end{tabular}}        & Avg    & {\color[HTML]{0000FF} {\ul 0.838}}    & {\color[HTML]{0000FF} {\ul 0.751}}    & 0.855                                 & 0.757                                 & 0.849                              & 0.754                                 & 0.853                                 & 0.755                                 & 0.847                                 & 0.753                                 & 0.848                                 & 0.753                                 & {\color[HTML]{FF0000} \textbf{0.827}} & {\color[HTML]{FF0000} \textbf{0.743}} & 0.872                                 & 0.766                                 & 1.969                              & 1.121                                 & 1.938                                 & 1.112                                 \\ \hline
                                                                                      & 96     & {\color[HTML]{FF0000} \textbf{1.086}} & {\color[HTML]{FF0000} \textbf{0.627}} & 1.169                                 & 0.652                                 & 1.186                              & 0.655                                 & {\color[HTML]{0000FF} {\ul 1.12}}     & {\color[HTML]{0000FF} {\ul 0.63}}     & 1.145                                 & 0.656                                 & 1.188                                 & 0.654                                 & 1.136                                 & 0.633                                 & 1.158                                 & 0.651                                 & 1.152                              & 0.661                                 & 1.506                                 & 0.782                                 \\
                                                                                      & 192    & {\color[HTML]{FF0000} \textbf{0.978}} & {\color[HTML]{FF0000} \textbf{0.627}} & 1.09                                  & 0.645                                 & 1.094                              & 0.651                                 & 1.048                                 & 0.641                                 & {\color[HTML]{0000FF} {\ul 1.034}}    & {\color[HTML]{0000FF} {\ul 0.629}}    & 1.089                                 & 0.646                                 & 1.075                                 & {\color[HTML]{0000FF} {\ul 0.629}}    & 1.092                                 & 0.641                                 & 1.069                              & 0.658                                 & 1.414                                 & 0.774                                 \\
                                                                                      & 336    & 1.021                                 & 0.648                                 & 1.092                                 & 0.658                                 & 1.065                              & 0.653                                 & {\color[HTML]{0000FF} {\ul 1.011}}    & {\color[HTML]{0000FF} {\ul 0.634}}    & {\color[HTML]{FF0000} \textbf{1.01}}  & 0.638                                 & 1.068                                 & 0.65                                  & 1.014                                 & {\color[HTML]{FF0000} \textbf{0.626}} & 1.022                                 & {\color[HTML]{FF0000} \textbf{0.626}} & 1.04                               & 0.654                                 & 2.097                                 & 0.782                                 \\
                                                                                      & 720    & {\color[HTML]{FF0000} \textbf{0.901}} & {\color[HTML]{FF0000} \textbf{0.619}} & 0.99                                  & 0.634                                 & 1.008                              & 0.635                                 & {\color[HTML]{0000FF} {\ul 0.959}}    & {\color[HTML]{FF0000} \textbf{0.619}} & 0.99                                  & 0.636                                 & 0.993                                 & 0.632                                 & 0.979                                 & {\color[HTML]{0000FF} {\ul 0.623}}    & 1.002                                 & 0.634                                 & 1.088                              & 0.681                                 & 1.492                                 & 0.821                                 \\ \cline{2-22} 
\multirow{-5}{*}{\begin{tabular}[c]{@{}c@{}}KDDCup\\ 2018\end{tabular}}               & Avg    & {\color[HTML]{FF0000} \textbf{0.997}} & {\color[HTML]{0000FF} {\ul 0.63}}     & 1.086                                 & 0.647                                 & 1.088                              & 0.648                                 & {\color[HTML]{0000FF} {\ul 1.035}}    & 0.631                                 & 1.045                                 & 0.64                                  & 1.085                                 & 0.646                                 & 1.051                                 & {\color[HTML]{FF0000} \textbf{0.628}} & 1.068                                 & 0.638                                 & 1.087                              & 0.664                                 & 1.627                                 & 0.79                                  \\ \hline
                                                                                      & 96     & 0.239                                 & 0.258                                 & 0.22                                  & 0.254                                 & 0.226                              & 0.246                                 & 0.225                                 & 0.248                                 & {\color[HTML]{0000FF} {\ul 0.219}}    & 0.245                                 & {\color[HTML]{FF0000} \textbf{0.216}} & {\color[HTML]{0000FF} {\ul 0.229}}    & {\color[HTML]{FF0000} \textbf{0.216}} & {\color[HTML]{FF0000} \textbf{0.222}} & 0.456                                 & 0.403                                 & 2.655                              & 1.093                                 & 1.995                                 & 0.954                                 \\
                                                                                      & 192    & 0.266                                 & 0.272                                 & 0.257                                 & 0.274                                 & {\color[HTML]{0000FF} {\ul 0.255}} & {\color[HTML]{0000FF} {\ul 0.255}}    & 0.259                                 & 0.267                                 & 0.377                                 & 0.367                                 & 0.258                                 & 0.261                                 & {\color[HTML]{FF0000} \textbf{0.244}} & {\color[HTML]{FF0000} \textbf{0.24}}  & 0.416                                 & 0.38                                  & 2.644                              & 1.1                                   & 2.029                                 & 0.964                                 \\
                                                                                      & 336    & 0.307                                 & 0.295                                 & 0.305                                 & 0.303                                 & 0.312                              & 0.306                                 & 0.297                                 & {\color[HTML]{0000FF} {\ul 0.284}}    & {\color[HTML]{FF0000} \textbf{0.289}} & 0.287                                 & 0.304                                 & 0.291                                 & {\color[HTML]{0000FF} {\ul 0.291}}    & {\color[HTML]{FF0000} \textbf{0.265}} & 0.451                                 & 0.4                                   & 2.663                              & 1.123                                 & 2.065                                 & 0.988                                 \\
                                                                                      & 720    & 0.381                                 & 0.332                                 & 0.384                                 & 0.341                                 & 0.386                              & 0.334                                 & 0.389                                 & 0.338                                 & {\color[HTML]{FF0000} \textbf{0.36}}  & {\color[HTML]{0000FF} {\ul 0.327}}    & 0.39                                  & 0.332                                 & {\color[HTML]{0000FF} {\ul 0.376}}    & {\color[HTML]{FF0000} \textbf{0.31}}  & 0.541                                 & 0.449                                 & 3.011                              & 1.208                                 & 2.238                                 & 1.052                                 \\ \cline{2-22} 
\multirow{-5}{*}{\begin{tabular}[c]{@{}c@{}}Pedestrian\\ Counts\end{tabular}}         & Avg    & 0.298                                 & 0.289                                 & {\color[HTML]{0000FF} {\ul 0.291}}    & 0.293                                 & 0.295                              & 0.285                                 & 0.292                                 & 0.284                                 & 0.311                                 & 0.307                                 & 0.292                                 & {\color[HTML]{0000FF} {\ul 0.278}}    & {\color[HTML]{FF0000} \textbf{0.282}} & {\color[HTML]{FF0000} \textbf{0.259}} & 0.466                                 & 0.408                                 & 2.744                              & 1.131                                 & 2.082                                 & 0.99                                  \\ \hline
Avg                                                                                   & Avg    & 0.77                                  & 0.484                                 & 0.691                                 & 0.458                                 & {\color[HTML]{0000FF} {\ul 0.547}} & {\color[HTML]{0000FF} {\ul 0.416}}    & 0.712                                 & 0.462                                 & 0.683                                 & 0.453                                 & {\color[HTML]{FF0000} \textbf{0.484}} & {\color[HTML]{FF0000} \textbf{0.407}} & 0.971                                 & 0.519                                 & 0.714                                 & 0.49                                  & 1.419                              & 0.726                                 & 1.392                                 & 0.746                                 \\ \hline
Rank                                                                                  & Avg    & 4.79                                  & 4.86                                  & {\color[HTML]{FF0000} \textbf{3.64}}  & {\color[HTML]{FF0000} \textbf{4.07}}  & 4.5                                & 4.64                                  & {\color[HTML]{0000FF} {\ul 4.21}}     & {\color[HTML]{0000FF} {\ul 4.29}}     & {\color[HTML]{0000FF} {\ul 4.21}}     & {\color[HTML]{0000FF} {\ul 4.29}}     & {\color[HTML]{0000FF} {\ul 4.21}}     & {\color[HTML]{0000FF} {\ul 4.29}}     & 5.57                                  & 5.14                                  & 6.64                                  & 6.14                                  & 8.36                               & 8.43                                  & 8.86                                  & 8.86                                  \\ \hline
\end{tabular}

}

\label{tab:full_results_mean}
\end{table*}

%% file: table_full_results_min.tex
\begin{table*}[t]
\tiny
\renewcommand{\arraystretch}{1.4}

\setlength{\tabcolsep}{2pt}
\caption{\textbf{Full results.}
\underline{Best values} for all prediction lengths. \textbf{\textcolor{red}{Best}} and {\textcolor{blue}{\underline{second-best}}} are highlighted.
}
\centering
\resizebox{\textwidth}{!}{
}             & Avg    & 4.583                                 & 1.182                                 & 3.745                                 & 1                                     & {\color[HTML]{0000FF} {\ul 1.496}}    & {\color[HTML]{0000FF} {\ul 0.338}} & 3.813                                 & 0.996                                 & 3.576                                 & 0.893                                 & {\color[HTML]{FF0000} \textbf{0.679}} & {\color[HTML]{FF0000} \textbf{0.225}} & 6.432                                 & 1.325                                 & 3.054                                 & 0.762                                 & 5.818                              & 1.321                              & 4.016                                 & 1.115                                 \\ \hline
                                                                                      & 96     & 0.769                                 & 0.65                                  & {\color[HTML]{0000FF} {\ul 0.662}}    & 0.595                                 & 0.67                                  & 0.597                              & 0.664                                 & {\color[HTML]{0000FF} {\ul 0.592}}    & {\color[HTML]{FF0000} \textbf{0.657}} & {\color[HTML]{FF0000} \textbf{0.591}} & {\color[HTML]{0000FF} {\ul 0.662}}    & {\color[HTML]{0000FF} {\ul 0.592}}    & 0.861                                 & 0.652                                 & 0.691                                 & 0.603                                 & 0.863                              & 0.675                              & 1.106                                 & 0.764                                 \\
                                                                                      & 192    & 1.002                                 & 0.746                                 & 0.833                                 & 0.669                                 & {\color[HTML]{0000FF} {\ul 0.828}}    & {\color[HTML]{0000FF} {\ul 0.665}} & 0.83                                  & 0.666                                 & {\color[HTML]{FF0000} \textbf{0.822}} & {\color[HTML]{FF0000} \textbf{0.663}} & {\color[HTML]{FF0000} \textbf{0.822}} & {\color[HTML]{FF0000} \textbf{0.663}} & 1.12                                  & 0.754                                 & 0.842                                 & 0.672                                 & 1.013                              & 0.739                              & 1.261                                 & 0.826                                 \\
                                                                                      & 336    & 1.286                                 & 0.846                                 & 1.047                                 & 0.747                                 & 1.044                                 & 0.745                              & 1.049                                 & 0.747                                 & {\color[HTML]{FF0000} \textbf{1.037}} & {\color[HTML]{0000FF} {\ul 0.743}}    & {\color[HTML]{0000FF} {\ul 1.038}}    & {\color[HTML]{FF0000} \textbf{0.742}} & 1.307                                 & 0.832                                 & 1.055                                 & 0.751                                 & 1.207                              & 0.809                              & 1.467                                 & 0.895                                 \\
                                                                                      & 720    & 1.544                                 & 0.963                                 & 1.361                                 & 0.879                                 & 1.355                                 & 0.875                              & 1.354                                 & 0.876                                 & {\color[HTML]{FF0000} \textbf{1.344}} & {\color[HTML]{FF0000} \textbf{0.871}} & {\color[HTML]{0000FF} {\ul 1.347}}    & {\color[HTML]{0000FF} {\ul 0.873}}    & 1.502                                 & 0.937                                 & {\color[HTML]{FF0000} \textbf{1.344}} & {\color[HTML]{FF0000} \textbf{0.871}} & 1.495                              & 0.92                               & 1.782                                 & 1.007                                 \\ \cline{2-22} 
\multirow{-5}{*}{TDBrain}                                                             & Avg    & 1.15                                  & 0.801                                 & 0.976                                 & 0.722                                 & 0.974                                 & 0.721                              & 0.974                                 & {\color[HTML]{0000FF} {\ul 0.72}}     & {\color[HTML]{FF0000} \textbf{0.965}} & {\color[HTML]{FF0000} \textbf{0.717}} & {\color[HTML]{0000FF} {\ul 0.967}}    & {\color[HTML]{FF0000} \textbf{0.717}} & 1.198                                 & 0.794                                 & 0.983                                 & 0.724                                 & 1.145                              & 0.786                              & 1.404                                 & 0.873                                 \\ \hline
                                                                                      & 96     & 0.528                                 & 0.44                                  & {\color[HTML]{0000FF} {\ul 0.518}}    & {\color[HTML]{0000FF} {\ul 0.428}}    & {\color[HTML]{FF0000} \textbf{0.515}} & 0.433                              & 0.528                                 & 0.441                                 & 0.522                                 & 0.43                                  & 0.535                                 & 0.437                                 & 0.528                                 & {\color[HTML]{FF0000} \textbf{0.406}} & 0.527                                 & 0.435                                 & 0.665                              & 0.504                              & 1.16                                  & 0.598                                 \\
                                                                                      & 192    & 0.569                                 & 0.463                                 & 0.572                                 & 0.454                                 & 0.568                                 & 0.458                              & 0.576                                 & 0.467                                 & 0.563                                 & {\color[HTML]{0000FF} {\ul 0.448}}    & {\color[HTML]{0000FF} {\ul 0.561}}    & 0.452                                 & {\color[HTML]{FF0000} \textbf{0.556}} & {\color[HTML]{FF0000} \textbf{0.421}} & 0.567                                 & 0.454                                 & 0.74                               & 0.541                              & 1.214                                 & 0.646                                 \\
                                                                                      & 336    & 0.591                                 & 0.473                                 & 0.59                                  & 0.466                                 & 0.593                                 & 0.467                              & 0.588                                 & 0.466                                 & 0.586                                 & 0.464                                 & {\color[HTML]{0000FF} {\ul 0.582}}    & {\color[HTML]{0000FF} {\ul 0.46}}     & {\color[HTML]{FF0000} \textbf{0.566}} & {\color[HTML]{FF0000} \textbf{0.433}} & 0.587                                 & 0.463                                 & 0.756                              & 0.548                              & 1.077                                 & 0.628                                 \\
                                                                                      & 720    & 0.641                                 & 0.507                                 & 0.623                                 & 0.486                                 & 0.612                                 & 0.487                              & 0.629                                 & 0.493                                 & 0.635                                 & 0.494                                 & {\color[HTML]{FF0000} \textbf{0.547}} & {\color[HTML]{FF0000} \textbf{0.445}} & {\color[HTML]{0000FF} {\ul 0.605}}    & {\color[HTML]{0000FF} {\ul 0.449}}    & 0.638                                 & 0.489                                 & 0.788                              & 0.571                              & 1.149                                 & 0.653                                 \\ \cline{2-22} 
\multirow{-5}{*}{BeijingAir}                                                          & Avg    & 0.582                                 & 0.471                                 & 0.576                                 & 0.458                                 & 0.572                                 & 0.461                              & 0.58                                  & 0.467                                 & 0.577                                 & 0.459                                 & {\color[HTML]{FF0000} \textbf{0.556}} & {\color[HTML]{0000FF} {\ul 0.449}}    & {\color[HTML]{0000FF} {\ul 0.564}}    & {\color[HTML]{FF0000} \textbf{0.427}} & 0.58                                  & 0.46                                  & 0.737                              & 0.541                              & 1.15                                  & 0.631                                 \\ \hline
                                                                                      & 96     & {\color[HTML]{0000FF} {\ul 0.007}}    & {\color[HTML]{0000FF} {\ul 0.014}}    & 0.009                                 & 0.039                                 & 0.009                                 & 0.048                              & {\color[HTML]{FF0000} \textbf{0.006}} & {\color[HTML]{FF0000} \textbf{0.013}} & 0.009                                 & 0.039                                 & {\color[HTML]{0000FF} {\ul 0.007}}    & 0.033                                 & 0.008                                 & {\color[HTML]{FF0000} \textbf{0.013}} & 0.279                                 & 0.388                                 & 0.931                              & 0.764                              & 1.264                                 & 0.862                                 \\
                                                                                      & 192    & {\color[HTML]{FF0000} \textbf{0.006}} & {\color[HTML]{FF0000} \textbf{0.013}} & 0.009                                 & 0.024                                 & {\color[HTML]{0000FF} {\ul 0.008}}    & 0.035                              & {\color[HTML]{FF0000} \textbf{0.006}} & {\color[HTML]{0000FF} {\ul 0.016}}    & 0.01                                  & 0.033                                 & {\color[HTML]{0000FF} {\ul 0.008}}    & 0.041                                 & {\color[HTML]{0000FF} {\ul 0.008}}    & 0.017                                 & 0.052                                 & 0.153                                 & 1.003                              & 0.81                               & 1.345                                 & 0.9                                   \\
                                                                                      & 336    & {\color[HTML]{0000FF} {\ul 0.008}}    & {\color[HTML]{0000FF} {\ul 0.017}}    & 0.009                                 & 0.04                                  & 0.013                                 & 0.055                              & {\color[HTML]{FF0000} \textbf{0.007}} & 0.02                                  & 0.011                                 & 0.03                                  & 0.01                                  & 0.048                                 & {\color[HTML]{0000FF} {\ul 0.008}}    & {\color[HTML]{FF0000} \textbf{0.011}} & 0.127                                 & 0.255                                 & 1.043                              & 0.829                              & 1.281                                 & 0.87                                  \\
                                                                                      & 720    & {\color[HTML]{FF0000} \textbf{0.011}} & {\color[HTML]{FF0000} \textbf{0.026}} & 0.015                                 & 0.05                                  & 0.014                                 & 0.053                              & {\color[HTML]{0000FF} {\ul 0.012}}    & {\color[HTML]{FF0000} \textbf{0.026}} & 0.014                                 & {\color[HTML]{0000FF} {\ul 0.028}}    & 0.013                                 & 0.04                                  & 0.014                                 & {\color[HTML]{FF0000} \textbf{0.026}} & 0.13                                  & 0.253                                 & 1.104                              & 0.856                              & 1.281                                 & 0.871                                 \\ \cline{2-22} 
\multirow{-5}{*}{\begin{tabular}[c]{@{}c@{}}Benzene\\ Concen-\\ tration\end{tabular}} & Avg    & {\color[HTML]{FF0000} \textbf{0.008}} & {\color[HTML]{0000FF} {\ul 0.018}}    & 0.011                                 & 0.038                                 & 0.011                                 & 0.048                              & {\color[HTML]{FF0000} \textbf{0.008}} & 0.019                                 & 0.011                                 & 0.032                                 & 0.01                                  & 0.041                                 & {\color[HTML]{0000FF} {\ul 0.009}}    & {\color[HTML]{FF0000} \textbf{0.017}} & 0.147                                 & 0.262                                 & 1.02                               & 0.815                              & 1.293                                 & 0.876                                 \\ \hline
                                                                                      & 96     & {\color[HTML]{0000FF} {\ul 0.805}}    & 0.729                                 & 0.814                                 & 0.731                                 & 0.808                                 & 0.727                              & 0.81                                  & 0.728                                 & 0.806                                 & 0.726                                 & {\color[HTML]{0000FF} {\ul 0.805}}    & {\color[HTML]{0000FF} {\ul 0.725}}    & {\color[HTML]{FF0000} \textbf{0.784}} & {\color[HTML]{FF0000} \textbf{0.715}} & 0.838                                 & 0.743                                 & 1.911                              & 1.092                              & 1.878                                 & 1.083                                 \\
                                                                                      & 192    & {\color[HTML]{0000FF} {\ul 0.839}}    & {\color[HTML]{0000FF} {\ul 0.75}}     & 0.855                                 & 0.756                                 & 0.849                                 & 0.754                              & 0.852                                 & 0.754                                 & 0.847                                 & 0.752                                 & 0.846                                 & 0.751                                 & {\color[HTML]{FF0000} \textbf{0.825}} & {\color[HTML]{FF0000} \textbf{0.742}} & 0.867                                 & 0.762                                 & 1.978                              & 1.122                              & 1.949                                 & 1.115                                 \\
                                                                                      & 336    & {\color[HTML]{0000FF} {\ul 0.85}}     & {\color[HTML]{0000FF} {\ul 0.759}}    & 0.864                                 & 0.763                                 & 0.862                                 & 0.762                              & 0.865                                 & 0.764                                 & 0.859                                 & 0.761                                 & 0.862                                 & 0.763                                 & {\color[HTML]{FF0000} \textbf{0.84}}  & {\color[HTML]{FF0000} \textbf{0.753}} & 0.883                                 & 0.774                                 & 1.976                              & 1.128                              & 1.944                                 & 1.119                                 \\
                                                                                      & 720    & {\color[HTML]{0000FF} {\ul 0.857}}    & {\color[HTML]{0000FF} {\ul 0.764}}    & 0.877                                 & 0.772                                 & 0.874                                 & 0.77                               & 0.874                                 & 0.77                                  & 0.871                                 & 0.769                                 & 0.872                                 & 0.769                                 & {\color[HTML]{FF0000} \textbf{0.852}} & {\color[HTML]{FF0000} \textbf{0.761}} & 0.892                                 & 0.779                                 & 2.011                              & 1.141                              & 1.979                                 & 1.132                                 \\ \cline{2-22} 
\multirow{-5}{*}{\begin{tabular}[c]{@{}c@{}}Australia\\ Rainfall\end{tabular}}        & Avg    & {\color[HTML]{0000FF} {\ul 0.838}}    & {\color[HTML]{0000FF} {\ul 0.751}}    & 0.852                                 & 0.756                                 & 0.848                                 & 0.753                              & 0.85                                  & 0.754                                 & 0.846                                 & 0.752                                 & 0.846                                 & 0.752                                 & {\color[HTML]{FF0000} \textbf{0.825}} & {\color[HTML]{FF0000} \textbf{0.743}} & 0.87                                  & 0.765                                 & 1.969                              & 1.121                              & 1.938                                 & 1.112                                 \\ \hline
                                                                                      & 96     & {\color[HTML]{FF0000} \textbf{1.082}} & {\color[HTML]{FF0000} \textbf{0.627}} & 1.14                                  & 0.647                                 & 1.183                                 & 0.654                              & 1.113                                 & {\color[HTML]{FF0000} \textbf{0.627}} & {\color[HTML]{0000FF} {\ul 1.111}}    & 0.645                                 & 1.185                                 & 0.653                                 & 1.126                                 & {\color[HTML]{0000FF} {\ul 0.628}}    & 1.141                                 & 0.65                                  & 1.152                              & 0.661                              & 1.506                                 & 0.782                                 \\
                                                                                      & 192    & {\color[HTML]{FF0000} \textbf{0.978}} & {\color[HTML]{FF0000} \textbf{0.626}} & 1.09                                  & 0.645                                 & 1.087                                 & 0.646                              & 1.044                                 & 0.636                                 & {\color[HTML]{0000FF} {\ul 1.034}}    & {\color[HTML]{0000FF} {\ul 0.628}}    & 1.084                                 & 0.645                                 & 1.072                                 & {\color[HTML]{FF0000} \textbf{0.626}} & 1.086                                 & 0.638                                 & 1.069                              & 0.658                              & 1.414                                 & 0.774                                 \\
                                                                                      & 336    & 1.006                                 & 0.641                                 & 1.085                                 & 0.654                                 & 1.059                                 & 0.65                               & 1.008                                 & {\color[HTML]{0000FF} {\ul 0.629}}    & {\color[HTML]{FF0000} \textbf{0.999}} & 0.63                                  & 1.034                                 & 0.635                                 & {\color[HTML]{0000FF} {\ul 1}}        & {\color[HTML]{FF0000} \textbf{0.621}} & 1.021                                 & {\color[HTML]{FF0000} \textbf{0.621}} & 1.04                               & 0.654                              & 2.097                                 & 0.782                                 \\
                                                                                      & 720    & {\color[HTML]{FF0000} \textbf{0.889}} & {\color[HTML]{FF0000} \textbf{0.616}} & 0.988                                 & 0.631                                 & 1.007                                 & 0.635                              & {\color[HTML]{0000FF} {\ul 0.956}}    & {\color[HTML]{0000FF} {\ul 0.617}}    & 0.975                                 & 0.629                                 & 0.99                                  & 0.632                                 & 0.977                                 & 0.622                                 & 1                                     & 0.634                                 & 1.088                              & 0.681                              & 1.492                                 & 0.821                                 \\ \cline{2-22} 
\multirow{-5}{*}{\begin{tabular}[c]{@{}c@{}}KDDCup\\ 2018\end{tabular}}               & Avg    & {\color[HTML]{FF0000} \textbf{0.989}} & {\color[HTML]{0000FF} {\ul 0.627}}    & 1.076                                 & 0.644                                 & 1.084                                 & 0.646                              & {\color[HTML]{0000FF} {\ul 1.03}}     & {\color[HTML]{0000FF} {\ul 0.627}}    & {\color[HTML]{0000FF} {\ul 1.03}}     & 0.633                                 & 1.073                                 & 0.641                                 & 1.044                                 & {\color[HTML]{FF0000} \textbf{0.624}} & 1.062                                 & 0.636                                 & 1.087                              & 0.664                              & 1.627                                 & 0.79                                  \\ \hline
                                                                                      & 96     & 0.238                                 & 0.257                                 & 0.218                                 & 0.252                                 & 0.22                                  & 0.241                              & 0.219                                 & 0.239                                 & {\color[HTML]{0000FF} {\ul 0.216}}    & 0.241                                 & {\color[HTML]{FF0000} \textbf{0.212}} & {\color[HTML]{0000FF} {\ul 0.227}}    & {\color[HTML]{0000FF} {\ul 0.216}}    & {\color[HTML]{FF0000} \textbf{0.222}} & 0.455                                 & 0.401                                 & 2.655                              & 1.093                              & 1.995                                 & 0.954                                 \\
                                                                                      & 192    & 0.265                                 & 0.272                                 & 0.255                                 & 0.273                                 & {\color[HTML]{0000FF} {\ul 0.253}}    & {\color[HTML]{0000FF} {\ul 0.255}} & 0.255                                 & 0.266                                 & 0.322                                 & 0.331                                 & {\color[HTML]{0000FF} {\ul 0.253}}    & 0.258                                 & {\color[HTML]{FF0000} \textbf{0.244}} & {\color[HTML]{FF0000} \textbf{0.24}}  & 0.415                                 & 0.379                                 & 2.644                              & 1.1                                & 2.029                                 & 0.964                                 \\
                                                                                      & 336    & 0.307                                 & 0.295                                 & 0.303                                 & 0.301                                 & 0.309                                 & 0.302                              & 0.295                                 & 0.283                                 & {\color[HTML]{FF0000} \textbf{0.284}} & {\color[HTML]{0000FF} {\ul 0.282}}    & 0.302                                 & 0.291                                 & {\color[HTML]{0000FF} {\ul 0.291}}    & {\color[HTML]{FF0000} \textbf{0.265}} & 0.451                                 & 0.399                                 & 2.663                              & 1.123                              & 2.065                                 & 0.988                                 \\
                                                                                      & 720    & 0.381                                 & 0.332                                 & 0.378                                 & 0.339                                 & 0.383                                 & 0.332                              & 0.375                                 & 0.328                                 & {\color[HTML]{FF0000} \textbf{0.358}} & {\color[HTML]{0000FF} {\ul 0.324}}    & 0.385                                 & 0.33                                  & {\color[HTML]{0000FF} {\ul 0.372}}    & {\color[HTML]{FF0000} \textbf{0.307}} & 0.54                                  & 0.448                                 & 3.011                              & 1.208                              & 2.238                                 & 1.052                                 \\ \cline{2-22} 
\multirow{-5}{*}{\begin{tabular}[c]{@{}c@{}}Pedestrian\\ Counts\end{tabular}}         & Avg    & 0.298                                 & 0.289                                 & 0.289                                 & 0.291                                 & 0.291                                 & 0.283                              & {\color[HTML]{0000FF} {\ul 0.286}}    & 0.279                                 & 0.295                                 & 0.295                                 & 0.288                                 & {\color[HTML]{0000FF} {\ul 0.277}}    & {\color[HTML]{FF0000} \textbf{0.281}} & {\color[HTML]{FF0000} \textbf{0.259}} & 0.465                                 & 0.407                                 & 2.744                              & 1.131                              & 2.082                                 & 0.99                                  \\ \hline
Avg                                                                                   & Avg    & 0.759                                 & 0.478                                 & 0.686                                 & 0.455                                 & {\color[HTML]{0000FF} {\ul 0.529}}    & {\color[HTML]{0000FF} {\ul 0.41}}  & 0.691                                 & 0.452                                 & 0.673                                 & 0.448                                 & {\color[HTML]{FF0000} \textbf{0.474}} & {\color[HTML]{FF0000} \textbf{0.403}} & 0.945                                 & 0.508                                 & 0.706                                 & 0.486                                 & 1.419                              & 0.726                              & 1.392                                 & 0.746                                 \\ \hline
Rank                                                                                  & Avg    & 4.64                                  & 4.71                                  & {\color[HTML]{FF0000} \textbf{3.86}}  & 4.29                                  & 4.5                                   & 5                                  & {\color[HTML]{FF0000} \textbf{3.86}}  & {\color[HTML]{FF0000} \textbf{3.79}}  & {\color[HTML]{0000FF} {\ul 4.29}}     & {\color[HTML]{0000FF} {\ul 4.14}}     & 4.43                                  & 4.43                                  & 5.43                                  & 4.86                                  & 6.57                                  & 6.36                                  & 8.5                                & 8.5                                & 8.93                                  & 8.93                                  \\ \hline
\end{tabular}
}

\label{tab:full_results_min}
\end{table*}

%% file: table_full_results_ipatch.tex
\begin{table*}[t]
\caption{\textbf{Full results iPatch.} We present TSLib (left) and UTSD datasets (right). See \Cref{tab:ipatch_champ} for average MSE, MAE, and Rank.
}
\centering
\resizebox{\textwidth}{!}{

\begin{tabular}{c|c|cccc|c|c|cccc}
\toprule
\multirow{3}{*}{Dataset}     & Model     & \multicolumn{4}{c|}{iPatch}                                    & \multirow{3}{*}{Dataset}              & Model     & \multicolumn{4}{c}{iPatch}                        \\ \cmidrule{2-6} \cmidrule{8-12} 
                             & Metric    & \multicolumn{2}{c}{MAE} & \multicolumn{2}{c|}{MSE}             &                                       & Metric    & \multicolumn{2}{c}{MAE} & \multicolumn{2}{c}{MSE} \\ \cmidrule(lr){2-2}\cmidrule(lr){3-4} \cmidrule(lr){5-6}\cmidrule(lr){8-8}\cmidrule(lr){9-10} \cmidrule(lr){11-12} 
                             & Statistic & Mean       & Min        & Mean                        & Min    &                                       & Statistic & Mean       & Min        & Mean       & Min        \\ \midrule
\multirow{5}{*}{ETTh1}       & 96        & 0.4074     & 0.3995     & 0.3877                      & 0.379  & \multirow{5}{*}{MotorImagery}         & 96        & 0.1433     & 0.1287     & 0.4238     & 0.3541     \\
                             & 192       & 0.4241     & 0.4221     & 0.415                       & 0.413  &                                       & 192       & 0.2633     & 0.2079     & 1.0616     & 0.7352     \\
                             & 336       & 0.4301     & 0.4284     & 0.4241                      & 0.4198 &                                       & 336       & 0.3999     & 0.3458     & 2.183      & 1.8659     \\
                             & 720       & 0.4635     & 0.4599     & 0.45                        & 0.4451 &                                       & 720       & 1.1959     & 1.1879     & 4.8701     & 4.8506     \\ \cmidrule{2-6} \cmidrule{8-12} 
                             & Average   & 0.4313     & 0.4275     & 0.4192                      & 0.4142 &                                       & Average   & 0.5006     & 0.4676     & 2.1346     & 1.9515     \\ \midrule
\multirow{5}{*}{ETTm1}       & 96        & 0.383      & 0.3726     & 0.3229                      & 0.3115 & \multirow{5}{*}{TDBrain}              & 96        & 0.5985     & 0.5909     & 0.6747     & 0.6592     \\
                             & 192       & 0.418      & 0.4171     & 0.3749                      & 0.3719 &                                       & 192       & 0.6704     & 0.665      & 0.8383     & 0.824      \\
                             & 336       & 0.4478     & 0.446      & 0.4289                      & 0.4259 &                                       & 336       & 0.7483     & 0.7423     & 1.0502     & 1.036      \\
                             & 720       & 0.4963     & 0.4919     & 0.4937                      & 0.4862 &                                       & 720       & 0.871      & 0.8704     & 1.3459     & 1.3451     \\ \cmidrule{2-6} \cmidrule{8-12} 
                             & Average   & 0.4363     & 0.4319     & 0.4051                      & 0.3989 &                                       & Average   & 0.722      & 0.7171     & 0.9773     & 0.9661     \\ \midrule
\multirow{5}{*}{ETTh2}       & 96        & 0.3445     & 0.3423     & 0.2935                      & 0.2902 & \multirow{5}{*}{BeijingAir}           & 96        & 0.4345     & 0.4302     & 0.5334     & 0.5231     \\
                             & 192       & 0.3953     & 0.3931     & 0.3759                      & 0.3711 &                                       & 192       & 0.4559     & 0.455      & 0.569      & 0.5643     \\
                             & 336       & 0.4337     & 0.4323     & 0.4226                      & 0.4205 &                                       & 336       & 0.4665     & 0.4658     & 0.5962     & 0.5941     \\
                             & 720       & 0.4443     & 0.4419     & 0.4197                      & 0.4172 &                                       & 720       & 0.4721     & 0.4685     & 0.5909     & 0.5843     \\ \cmidrule{2-6} \cmidrule{8-12} 
                             & Average   & 0.4045     & 0.4024     & 0.3779                      & 0.3747 &                                       & Average   & 0.4573     & 0.4549     & 0.5724     & 0.5665     \\ \midrule
\multirow{5}{*}{ETTm2}       & 96        & 0.2282     & 0.2281     & 0.1145                      & 0.1134 & \multirow{5}{*}{BenzeneConcentration} & 96        & 0.0197     & 0.0192     & 0.0059     & 0.0059     \\
                             & 192       & 0.2595     & 0.2593     & 0.146                       & 0.1457 &                                       & 192       & 0.0247     & 0.0244     & 0.0112     & 0.0111     \\
                             & 336       & 0.2865     & 0.285      & 0.18                        & 0.1786 &                                       & 336       & 0.0188     & 0.0166     & 0.0077     & 0.0076     \\
                             & 720       & 0.3181     & 0.3148     & 0.214                       & 0.2099 &                                       & 720       & 0.0315     & 0.0302     & 0.0152     & 0.0149     \\ \cmidrule{2-6} \cmidrule{8-12} 
                             & Average   & 0.2731     & 0.2718     & \multicolumn{1}{c|}{0.1636} & 0.1619 &                                       & Average   & 0.0237     & 0.0226     & 0.01       & 0.0098     \\ \midrule
\multirow{5}{*}{Electricity} & 96        & 0.2315     & 0.2304     & 0.1333                      & 0.133  & \multirow{5}{*}{AustraliaRainfall}    & 96        & 0.7284     & 0.7274     & 0.8115     & 0.8093     \\
                             & 192       & 0.2471     & 0.2456     & 0.1511                      & 0.15   &                                       & 192       & 0.7533     & 0.7524     & 0.85       & 0.8484     \\
                             & 336       & 0.2808     & 0.2782     & 0.1798                      & 0.1769 &                                       & 336       & 0.7623     & 0.7608     & 0.8624     & 0.8583     \\
                             & 720       & 0.3196     & 0.3159     & 0.2295                      & 0.2268 &                                       & 720       & 0.7691     & 0.7683     & 0.8719     & 0.8702     \\ \cmidrule{2-6} \cmidrule{8-12} 
                             & Average   & 0.2698     & 0.2675     & 0.1734                      & 0.1717 &                                       & Average   & 0.7533     & 0.7522     & 0.8489     & 0.8466     \\ \midrule
\multirow{5}{*}{Weather}     & 96        & 0.204      & 0.2031     & 0.1532                      & 0.1525 & \multirow{5}{*}{KDDCup2018}           & 96        & 0.6556     & 0.6539     & 1.1853     & 1.1828     \\
                             & 192       & 0.2501     & 0.2494     & 0.2032                      & 0.2016 &                                       & 192       & 0.6455     & 0.6454     & 1.0951     & 1.0948     \\
                             & 336       & 0.2879     & 0.2878     & 0.2516                      & 0.2511 &                                       & 336       & 0.6397     & 0.6366     & 1.0313     & 1.0267     \\
                             & 720       & 0.3348     & 0.3337     & 0.3202                      & 0.3192 &                                       & 720       & 0.6504     & 0.6478     & 1.0057     & 0.9874     \\ \cmidrule{2-6} \cmidrule{8-12} 
                             & Average   & 0.2692     & 0.2685     & 0.232                       & 0.2311 &                                       & Average   & 0.6478     & 0.6459     & 1.0793     & 1.0729     \\ \midrule
\multirow{5}{*}{Exchange}    & 96        & 0.2146     & 0.2114     & 0.0934                      & 0.0906 & \multirow{5}{*}{PedestrianCounts}     & 96        & 0.2316     & 0.2309     & 0.2139     & 0.213      \\
                             & 192       & 0.3127     & 0.3067     & 0.1901                      & 0.1853 &                                       & 192       & 0.2628     & 0.2498     & 0.2564     & 0.2475     \\
                             & 336       & 0.4260     & 0.4219     & 0.3461                      & 0.3397 &                                       & 336       & 0.2797     & 0.274      & 0.2939     & 0.2894     \\
                             & 720       & 0.8205     & 0.8104     & 1.1578                      & 1.1318 &                                       & 720       & 0.3196     & 0.318      & 0.3726     & 0.372      \\ \cmidrule{2-6} \cmidrule{8-12} 
                             & Average   & 0.4435     & 0.4376     & 0.4469                      & 0.4369  &                                       & Average   & 0.2734     & 0.2682     & 0.2842     & 0.2805    \\
\bottomrule
\end{tabular}

}

\label{tab:ipatch_results}
\end{table*}

%% file: main.bbl
\begin{thebibliography}{79}
\providecommand{\natexlab}[1]{#1}
\providecommand{\url}[1]{\texttt{#1}}
\expandafter\ifx\csname urlstyle\endcsname\relax
  \providecommand{\doi}[1]{doi: #1}\else
  \providecommand{\doi}{doi: \begingroup \urlstyle{rm}\Url}\fi

\bibitem[Akiba et~al.(2019)Akiba, Sano, Yanase, Ohta, and Koyama]{optuna_2019}
Takuya Akiba, Shotaro Sano, Toshihiko Yanase, Takeru Ohta, and Masanori Koyama.
\newblock Optuna: A next-generation hyperparameter optimization framework.
\newblock In \emph{Proceedings of the 25th {ACM} {SIGKDD} International Conference on Knowledge Discovery and Data Mining}, 2019.

\bibitem[Aksu et~al.(2024)Aksu, Woo, Liu, Liu, Liu, Savarese, Xiong, and Sahoo]{aksu2024gift}
Taha Aksu, Gerald Woo, Juncheng Liu, Xu~Liu, Chenghao Liu, Silvio Savarese, Caiming Xiong, and Doyen Sahoo.
\newblock Gift-eval: A benchmark for general time series forecasting model evaluation.
\newblock \emph{arXiv preprint arXiv:2410.10393}, 2024.

\bibitem[Alharthi \& Mahmood(2024)Alharthi and Mahmood]{alharthi2024xlstmtime}
Musleh Alharthi and Ausif Mahmood.
\newblock xlstmtime: Long-term time series forecasting with xlstm.
\newblock \emph{AI}, 5\penalty0 (3):\penalty0 1482--1495, 2024.

\bibitem[Ansari et~al.(2024)Ansari, Stella, Turkmen, Zhang, Mercado, Shen, Shchur, Rangapuram, Arango, Kapoor, Zschiegner, Maddix, Wang, Mahoney, Torkkola, Wilson, Bohlke-Schneider, and Wang]{ansari2024chronos}
Abdul~Fatir Ansari, Lorenzo Stella, Ali~Caner Turkmen, Xiyuan Zhang, Pedro Mercado, Huibin Shen, Oleksandr Shchur, Syama~Sundar Rangapuram, Sebastian~Pineda Arango, Shubham Kapoor, Jasper Zschiegner, Danielle~C. Maddix, Hao Wang, Michael~W. Mahoney, Kari Torkkola, Andrew~Gordon Wilson, Michael Bohlke-Schneider, and Bernie Wang.
\newblock Chronos: Learning the language of time series.
\newblock \emph{Transactions on Machine Learning Research}, 2024.
\newblock ISSN 2835-8856.

\bibitem[Bechler-Speicher et~al.(2025)Bechler-Speicher, Finkelshtein, Frasca, Müller, Tönshoff, Siraudin, Zaverkin, Bronstein, Niepert, Perozzi, Galkin, and Morris]{bechler2025_graphbenchmarking}
Maya Bechler-Speicher, Ben Finkelshtein, Fabrizio Frasca, Luis Müller, Jan Tönshoff, Antoine Siraudin, Viktor Zaverkin, Michael~M. Bronstein, Mathias Niepert, Bryan Perozzi, Mikhail Galkin, and Christopher Morris.
\newblock Position: {Graph} {Learning} {Will} {Lose} {Relevance} {Due} {To} {Poor} {Benchmarks}, February 2025.
\newblock URL \url{http://arxiv.org/abs/2502.14546}.
\newblock arXiv:2502.14546 [cs].

\bibitem[Beck et~al.(2024)Beck, P{\"o}ppel, Spanring, Auer, Prudnikova, Kopp, Klambauer, Brandstetter, and Hochreiter]{beck2024xlstm}
Maximilian Beck, Korbinian P{\"o}ppel, Markus Spanring, Andreas Auer, Oleksandra Prudnikova, Michael Kopp, G{\"u}nter Klambauer, Johannes Brandstetter, and Sepp Hochreiter.
\newblock xlstm: Extended long short-term memory.
\newblock \emph{Advances in Neural Information Processing Systems}, 2024.

\bibitem[Bergmeir(2024)]{bergmeir2024llms}
Christoph Bergmeir.
\newblock Llms and foundational models: Not (yet) as good as hoped.
\newblock \emph{Foresight: The International Journal of Applied Forecasting}, 73:\penalty0 33--38, 2024.

\bibitem[Box \& Pierce(1970)Box and Pierce]{BoxPierce1970}
George E.~P. Box and David~A. Pierce.
\newblock Distribution of residual autocorrelations in autoregressive-integrated moving average time series models.
\newblock \emph{Journal of the American Statistical Association}, 65\penalty0 (332):\penalty0 1509--1526, 1970.

\bibitem[Breiman(2001)]{breiman2001random}
Leo Breiman.
\newblock Random forests.
\newblock \emph{Machine learning}, 45:\penalty0 5--32, 2001.
\newblock \doi{10.1023/A:1010933404324}.

\bibitem[Brigato et~al.(2021)Brigato, Barz, Iocchi, and Denzler]{Brigato_2021_ICCV}
Lorenzo Brigato, Bj\"orn Barz, Luca Iocchi, and Joachim Denzler.
\newblock Tune it or don't use it: Benchmarking data-efficient image classification.
\newblock In \emph{Proceedings of the IEEE/CVF International Conference on Computer Vision (ICCV) Workshops}, pp.\  1071--1080, October 2021.

\bibitem[Cao et~al.(2024)Cao, Jia, Arik, Pfister, Zheng, Ye, and Liu]{cao2023tempo}
Defu Cao, Furong Jia, Sercan~O Arik, Tomas Pfister, Yixiang Zheng, Wen Ye, and Yan Liu.
\newblock Tempo: Prompt-based generative pre-trained transformer for time series forecasting.
\newblock In \emph{The Twelfth International Conference on Learning Representations}, 2024.

\bibitem[Challu et~al.(2023)Challu, Olivares, Oreshkin, Ramirez, Canseco, and Dubrawski]{challu2023nhits}
Cristian Challu, Kin~G Olivares, Boris~N Oreshkin, Federico~Garza Ramirez, Max~Mergenthaler Canseco, and Artur Dubrawski.
\newblock Nhits: Neural hierarchical interpolation for time series forecasting.
\newblock In \emph{Proceedings of the AAAI Conference on Artificial Intelligence}, pp.\  6989--6997, 2023.

\bibitem[Chang et~al.(2025)Chang, Wang, Peng, and Chen]{chang2023llm4ts}
Ching Chang, Wei-Yao Wang, Wen-Chih Peng, and Tien-Fu Chen.
\newblock Llm4ts: Aligning pre-trained llms as data-efficient time-series forecasters.
\newblock \emph{ACM Trans. Intell. Syst. Technol.}, February 2025.
\newblock ISSN 2157-6904.
\newblock \doi{10.1145/3719207}.
\newblock URL \url{https://doi.org/10.1145/3719207}.

\bibitem[Chen \& Guestrin(2016)Chen and Guestrin]{chen2016xgboost}
Tianqi Chen and Carlos Guestrin.
\newblock Xgboost: A scalable tree boosting system.
\newblock In \emph{Proceedings of the 22nd acm sigkdd international conference on knowledge discovery and data mining}, pp.\  785--794, 2016.

\bibitem[Chen et~al.(2025)Chen, C{\'e}spedes, and Barnaghi]{chencloser}
Yu~Chen, Nathalia C{\'e}spedes, and Payam Barnaghi.
\newblock A closer look at transformers for time series forecasting: Understanding why they work and where they struggle.
\newblock In \emph{Forty-second International Conference on Machine Learning}, 2025.

\bibitem[Das et~al.(2023)Das, Kong, Leach, Mathur, Sen, and Yu]{das_long-term_2024}
Abhimanyu Das, Weihao Kong, Andrew Leach, Shaan~K Mathur, Rajat Sen, and Rose Yu.
\newblock Long-term forecasting with tide: Time-series dense encoder.
\newblock \emph{Transactions on Machine Learning Research}, 2023.
\newblock ISSN 2835-8856.

\bibitem[Dem{\v{s}}ar(2006)]{demvsar2006statistical}
Janez Dem{\v{s}}ar.
\newblock Statistical comparisons of classifiers over multiple data sets.
\newblock \emph{The Journal of Machine learning research}, 7:\penalty0 1--30, 2006.

\bibitem[Ekambaram et~al.(2024)Ekambaram, Jati, Dayama, Mukherjee, Nguyen, Gifford, Reddy, and Kalagnanam]{ekambaram2024ttms}
Vijay Ekambaram, Arindam Jati, Pankaj Dayama, Sumanta Mukherjee, Nam~H Nguyen, Wesley~M. Gifford, Chandra Reddy, and Jayant Kalagnanam.
\newblock Tiny time mixers (ttms): Fast pre-trained models for enhanced zero/few-shot forecasting of multivariate time series.
\newblock In \emph{The Thirty-eighth Annual Conference on Neural Information Processing Systems}, 2024.

\bibitem[Eriksson et~al.(2025)Eriksson, Purificato, Noroozian, Vinagre, Chaslot, Gomez, and Fernandez-Llorca]{eriksson2025_benchmarking}
Maria Eriksson, Erasmo Purificato, Arman Noroozian, Joao Vinagre, Guillaume Chaslot, Emilia Gomez, and David Fernandez-Llorca.
\newblock Can {We} {Trust} {AI} {Benchmarks}? {An} {Interdisciplinary} {Review} of {Current} {Issues} in {AI} {Evaluation}, February 2025.
\newblock URL \url{http://arxiv.org/abs/2502.06559}.
\newblock arXiv:2502.06559 [cs].

\bibitem[Friedman(2001)]{friedman2001greedy}
Jerome~H Friedman.
\newblock Greedy function approximation: a gradient boosting machine.
\newblock \emph{Annals of statistics}, pp.\  1189--1232, 2001.

\bibitem[Friedman(1937)]{friedman1937use}
Milton Friedman.
\newblock The use of ranks to avoid the assumption of normality implicit in the analysis of variance.
\newblock \emph{Journal of the american statistical association}, 32\penalty0 (200):\penalty0 675--701, 1937.

\bibitem[Friedman(1940)]{friedman1940comparison}
Milton Friedman.
\newblock A comparison of alternative tests of significance for the problem of m rankings.
\newblock \emph{The annals of mathematical statistics}, 11\penalty0 (1):\penalty0 86--92, 1940.

\bibitem[Garza \& Mergenthaler-Canseco(2023)Garza and Mergenthaler-Canseco]{garza2023timegpt}
Azul Garza and Max Mergenthaler-Canseco.
\newblock Timegpt-1.
\newblock \emph{arXiv preprint arXiv:2310.03589}, 2023.

\bibitem[Gruver et~al.(2024)Gruver, Finzi, Qiu, and Wilson]{gruver2024large}
Nate Gruver, Marc Finzi, Shikai Qiu, and Andrew~G Wilson.
\newblock Large language models are zero-shot time series forecasters.
\newblock \emph{Advances in Neural Information Processing Systems}, 36, 2024.

\bibitem[Gu \& Dao(2024)Gu and Dao]{gu2024mamba}
Albert Gu and Tri Dao.
\newblock Mamba: Linear-time sequence modeling with selective state spaces.
\newblock In \emph{First conference on language modeling}, 2024.

\bibitem[Herrmann et~al.(2024)Herrmann, Lange, Eggensperger, Casalicchio, Wever, Feurer, R{\"u}gamer, H{\"u}llermeier, Boulesteix, and Bischl]{herrmann2024position}
Moritz Herrmann, F~Julian~D Lange, Katharina Eggensperger, Giuseppe Casalicchio, Marcel Wever, Matthias Feurer, David R{\"u}gamer, Eyke H{\"u}llermeier, Anne-Laure Boulesteix, and Bernd Bischl.
\newblock Position: Why we must rethink empirical research in machine learning.
\newblock In \emph{International Conference on Machine Learning}, pp.\  18228--18247. PMLR, 2024.

\bibitem[Hewamalage et~al.(2023)Hewamalage, Ackermann, and Bergmeir]{hewamalage_forecast_2023}
Hansika Hewamalage, Klaus Ackermann, and Christoph Bergmeir.
\newblock Forecast evaluation for data scientists: common pitfalls and best practices.
\newblock \emph{Data Mining and Knowledge Discovery}, 37\penalty0 (2):\penalty0 788--832, March 2023.
\newblock ISSN 1573-756X.
\newblock \doi{10.1007/s10618-022-00894-5}.
\newblock URL \url{https://doi.org/10.1007/s10618-022-00894-5}.

\bibitem[Hochreiter \& Schmidhuber(1997)Hochreiter and Schmidhuber]{hochreiter1997long}
Sepp Hochreiter and Jürgen Schmidhuber.
\newblock Long short-term memory.
\newblock \emph{Neural Computation}, 9\penalty0 (8):\penalty0 1735--1780, 1997.
\newblock \doi{10.1162/neco.1997.9.8.1735}.

\bibitem[Hyndman et~al.(2008)Hyndman, Koehler, Ord, and Snyder]{hyndman2008forecasting}
Rob Hyndman, Anne~B Koehler, J~Keith Ord, and Ralph~D Snyder.
\newblock \emph{Forecasting with exponential smoothing: the state space approach}.
\newblock Springer Science \& Business Media, 2008.

\bibitem[Jin et~al.(2024)Jin, Zhang, Chen, Zhang, Liang, Yang, Wang, Pan, and Wen]{jin_position_2024}
Ming Jin, Yifan Zhang, Wei Chen, Kexin Zhang, Yuxuan Liang, Bin Yang, Jindong Wang, Shirui Pan, and Qingsong Wen.
\newblock Position: {What} {Can} {Large} {Language} {Models} {Tell} {Us} about {Time} {Series} {Analysis}.
\newblock In \emph{Proceedings of the 41st {International} {Conference} on {Machine} {Learning}}, pp.\  22260--22276. PMLR, July 2024.
\newblock URL \url{https://proceedings.mlr.press/v235/jin24i.html}.
\newblock ISSN: 2640-3498.

\bibitem[Jordan et~al.(2024)Jordan, White, Da~Silva, White, and Thomas]{jordan2024position}
Scott~M Jordan, Adam White, Bruno~Castro Da~Silva, Martha White, and Philip~S Thomas.
\newblock Position: Benchmarking is limited in reinforcement learning research.
\newblock In \emph{International Conference on Machine Learning}, pp.\  22551--22569. PMLR, 2024.

\bibitem[Karaouli et~al.(2025)Karaouli, Coquenet, Fromont, Mermillod, and Reyboz]{karaouli2025foundational}
Nouha Karaouli, Denis Coquenet, Elisa Fromont, Martial Mermillod, and Marina Reyboz.
\newblock How foundational are foundation models for time series forecasting?
\newblock \emph{arXiv preprint arXiv:2510.00742}, 2025.

\bibitem[Ke et~al.(2017)Ke, Meng, Finley, Wang, Chen, Ma, Ye, and Liu]{ke2017lightgbm}
Guolin Ke, Qi~Meng, Thomas Finley, Taifeng Wang, Wei Chen, Weidong Ma, Qiwei Ye, and Tie-Yan Liu.
\newblock Lightgbm: A highly efficient gradient boosting decision tree.
\newblock \emph{Advances in neural information processing systems}, 30, 2017.

\bibitem[Kim et~al.(2024)Kim, Kim, Kim, Lee, and Yoon]{kim2024comprehensive}
Jongseon Kim, Hyungjoon Kim, HyunGi Kim, Dongjun Lee, and Sungroh Yoon.
\newblock A comprehensive survey of time series forecasting: Architectural diversity and open challenges.
\newblock \emph{arXiv preprint arXiv:2411.05793}, 2024.

\bibitem[Koopmans(1995)]{koopmans1995spectral}
Lambert~H Koopmans.
\newblock \emph{The spectral analysis of time series}.
\newblock Elsevier, 1995.
\newblock \doi{https://doi.org/10.1016/B978-0-12-419251-5.X5000-5}.

\bibitem[Kraus et~al.(2024)Kraus, Divo, Dhami, and Kersting]{kraus2024xlstm}
Maurice Kraus, Felix Divo, Devendra~Singh Dhami, and Kristian Kersting.
\newblock xlstm-mixer: Multivariate time series forecasting by mixing via scalar memories.
\newblock \emph{arXiv preprint arXiv:2410.16928}, 2024.

\bibitem[Li et~al.(2019)Li, Jin, Xuan, Zhou, Chen, Wang, and Yan]{li_enhancing_2019}
Shiyang Li, Xiaoyong Jin, Yao Xuan, Xiyou Zhou, Wenhu Chen, Yu-Xiang Wang, and Xifeng Yan.
\newblock Enhancing the {Locality} and {Breaking} the {Memory} {Bottleneck} of {Transformer} on {Time} {Series} {Forecasting}.
\newblock In \emph{Advances in {Neural} {Information} {Processing} {Systems}}, volume~32. Curran Associates, Inc., 2019.

\bibitem[Li et~al.(2023)Li, Qi, Li, and Xu]{li_rlinear_2023}
Zhe Li, Shiyi Qi, Yiduo Li, and Zenglin Xu.
\newblock Revisiting {Long}-term {Time} {Series} {Forecasting}: {An} {Investigation} on {Linear} {Mapping}, May 2023.
\newblock URL \url{http://arxiv.org/abs/2305.10721}.
\newblock arXiv:2305.10721 [cs].

\bibitem[Liu et~al.(2022{\natexlab{a}})Liu, Zeng, Chen, Xu, LAI, Ma, and Xu]{liu2022scinet}
Minhao Liu, Ailing Zeng, Muxi Chen, Zhijian Xu, Qiuxia LAI, Lingna Ma, and Qiang Xu.
\newblock Scinet: Time series modeling and forecasting with sample convolution and interaction.
\newblock In S.~Koyejo, S.~Mohamed, A.~Agarwal, D.~Belgrave, K.~Cho, and A.~Oh (eds.), \emph{Advances in Neural Information Processing Systems}, volume~35, pp.\  5816--5828. Curran Associates, Inc., 2022{\natexlab{a}}.

\bibitem[Liu et~al.(2022{\natexlab{b}})Liu, Yu, Liao, Li, Lin, Liu, and Dustdar]{liu2022pyraformer}
Shizhan Liu, Hang Yu, Cong Liao, Jianguo Li, Weiyao Lin, Alex~X. Liu, and Schahram Dustdar.
\newblock Pyraformer: Low-complexity pyramidal attention for long-range time series modeling and forecasting.
\newblock In \emph{International Conference on Learning Representations}, 2022{\natexlab{b}}.
\newblock URL \url{https://openreview.net/forum?id=0EXmFzUn5I}.

\bibitem[Liu et~al.(2022{\natexlab{c}})Liu, Wu, Wang, and Long]{liu2022stationary}
Yong Liu, Haixu Wu, Jianmin Wang, and Mingsheng Long.
\newblock Non-stationary transformers: Exploring the stationarity in time series forecasting.
\newblock In S.~Koyejo, S.~Mohamed, A.~Agarwal, D.~Belgrave, K.~Cho, and A.~Oh (eds.), \emph{Advances in Neural Information Processing Systems}, volume~35, pp.\  9881--9893. Curran Associates, Inc., 2022{\natexlab{c}}.

\bibitem[Liu et~al.(2024{\natexlab{a}})Liu, Hu, Zhang, Wu, Wang, Ma, and Long]{liu_itransformer_2024}
Yong Liu, Tengge Hu, Haoran Zhang, Haixu Wu, Shiyu Wang, Lintao Ma, and Mingsheng Long.
\newblock itransformer: Inverted transformers are effective for time series forecasting.
\newblock In \emph{The Twelfth International Conference on Learning Representations}, 2024{\natexlab{a}}.

\bibitem[Liu et~al.(2024{\natexlab{b}})Liu, Zhang, Li, Huang, Wang, and Long]{liu2024timer}
Yong Liu, Haoran Zhang, Chenyu Li, Xiangdong Huang, Jianmin Wang, and Mingsheng Long.
\newblock Timer: Generative pre-trained transformers are large time series models.
\newblock In \emph{Forty-first International Conference on Machine Learning}, 2024{\natexlab{b}}.
\newblock URL \url{https://openreview.net/forum?id=bYRYb7DMNo}.

\bibitem[Luo \& Wang(2024)Luo and Wang]{luo2024moderntcn}
Donghao Luo and Xue Wang.
\newblock Moderntcn: A modern pure convolution structure for general time series analysis.
\newblock In \emph{The twelfth international conference on learning representations}, pp.\  1--43, 2024.

\bibitem[McElfresh et~al.(2024)McElfresh, Khandagale, Valverde, Prasad~C, Ramakrishnan, Goldblum, and White]{mcelfresh2024neural}
Duncan McElfresh, Sujay Khandagale, Jonathan Valverde, Vishak Prasad~C, Ganesh Ramakrishnan, Micah Goldblum, and Colin White.
\newblock When do neural nets outperform boosted trees on tabular data?
\newblock \emph{Advances in Neural Information Processing Systems}, 36, 2024.

\bibitem[McIntosh et~al.(2025)McIntosh, Susnjak, Arachchilage, Liu, Xu, Watters, and Halgamuge]{mcintosh2024_llmbenchmarking}
Timothy~R McIntosh, Teo Susnjak, Nalin Arachchilage, Tong Liu, Dan Xu, Paul Watters, and Malka~N Halgamuge.
\newblock Inadequacies of large language model benchmarks in the era of generative artificial intelligence.
\newblock \emph{IEEE Transactions on Artificial Intelligence}, 2025.

\bibitem[Nie et~al.(2023)Nie, Nguyen, Sinthong, and Kalagnanam]{nie_time_2023}
Yuqi Nie, Nam~H Nguyen, Phanwadee Sinthong, and Jayant Kalagnanam.
\newblock A time series is worth 64 words: Long-term forecasting with transformers.
\newblock In \emph{The Eleventh International Conference on Learning Representations}, 2023.

\bibitem[Oreshkin et~al.(2020)Oreshkin, Carpov, Chapados, and Bengio]{Oreshkin2020}
Boris~N. Oreshkin, Dmitri Carpov, Nicolas Chapados, and Yoshua Bengio.
\newblock N-beats: Neural basis expansion analysis for interpretable time series forecasting.
\newblock In \emph{International Conference on Learning Representations}, 2020.
\newblock URL \url{https://openreview.net/forum?id=r1ecqn4YwB}.

\bibitem[Qiu et~al.(2024)Qiu, Hu, Zhou, Wu, Du, Zhang, Guo, Zhou, Jensen, Sheng, and Yang]{qiu2024tfb}
Xiangfei Qiu, Jilin Hu, Lekui Zhou, Xingjian Wu, Junyang Du, Buang Zhang, Chenjuan Guo, Aoying Zhou, Christian~S. Jensen, Zhenli Sheng, and Bin Yang.
\newblock Tfb: Towards comprehensive and fair benchmarking of time series forecasting methods.
\newblock \emph{Proc. VLDB Endow.}, 17\penalty0 (9):\penalty0 2363–2377, May 2024.
\newblock ISSN 2150-8097.
\newblock \doi{10.14778/3665844.3665863}.
\newblock URL \url{https://doi.org/10.14778/3665844.3665863}.

\bibitem[Rasul et~al.(2023)Rasul, Ashok, Williams, Khorasani, Adamopoulos, Bhagwatkar, Bilo{\v{s}}, Ghonia, Hassen, Schneider, Garg, Drouin, Chapados, Nevmyvaka, and Rish]{rasul2023lag}
Kashif Rasul, Arjun Ashok, Andrew~Robert Williams, Arian Khorasani, George Adamopoulos, Rishika Bhagwatkar, Marin Bilo{\v{s}}, Hena Ghonia, Nadhir Hassen, Anderson Schneider, Sahil Garg, Alexandre Drouin, Nicolas Chapados, Yuriy Nevmyvaka, and Irina Rish.
\newblock Lag-llama: Towards foundation models for time series forecasting.
\newblock In \emph{R0-FoMo:Robustness of Few-shot and Zero-shot Learning in Large Foundation Models}, 2023.

\bibitem[Salzberg(1997)]{salzberg1997comparing}
Steven~L Salzberg.
\newblock On comparing classifiers: Pitfalls to avoid and a recommended approach.
\newblock \emph{Data mining and knowledge discovery}, 1:\penalty0 317--328, 1997.

\bibitem[Sarfraz et~al.(2024)Sarfraz, Chen, Layer, Peng, and Koulakis]{sarfraz24a_quo_vadis}
M.~Saquib Sarfraz, Mei-Yen Chen, Lukas Layer, Kunyu Peng, and Marios Koulakis.
\newblock Position: Quo vadis, unsupervised time series anomaly detection?
\newblock In Ruslan Salakhutdinov, Zico Kolter, Katherine Heller, Adrian Weller, Nuria Oliver, Jonathan Scarlett, and Felix Berkenkamp (eds.), \emph{Proceedings of the 41st International Conference on Machine Learning}, volume 235 of \emph{Proceedings of Machine Learning Research}, pp.\  43461--43476. PMLR, 21--27 Jul 2024.
\newblock URL \url{https://proceedings.mlr.press/v235/sarfraz24a.html}.

\bibitem[Sezer et~al.(2020)Sezer, Gudelek, and Ozbayoglu]{SEZER2020106181}
Omer~Berat Sezer, Mehmet~Ugur Gudelek, and Ahmet~Murat Ozbayoglu.
\newblock Financial time series forecasting with deep learning : A systematic literature review: 2005–2019.
\newblock \emph{Applied Soft Computing}, 90:\penalty0 106181, 2020.
\newblock ISSN 1568-4946.
\newblock \doi{https://doi.org/10.1016/j.asoc.2020.106181}.
\newblock URL \url{https://www.sciencedirect.com/science/article/pii/S1568494620301216}.

\bibitem[Shao et~al.(2024)Shao, Wang, Xu, Wei, Yu, Zhang, Yao, Sun, Jin, Cao, et~al.]{shao2024exploring}
Zezhi Shao, Fei Wang, Yongjun Xu, Wei Wei, Chengqing Yu, Zhao Zhang, Di~Yao, Tao Sun, Guangyin Jin, Xin Cao, et~al.
\newblock Exploring progress in multivariate time series forecasting: Comprehensive benchmarking and heterogeneity analysis.
\newblock \emph{IEEE Transactions on Knowledge and Data Engineering}, 2024.

\bibitem[Shi et~al.(2024)Shi, Ma, Ma, and Li]{ShiMML24}
Jingzhe Shi, Qinwei Ma, Huan Ma, and Lei Li.
\newblock Scaling law for time series forecasting.
\newblock In Amir Globersons, Lester Mackey, Danielle Belgrave, Angela Fan, Ulrich Paquet, Jakub~M. Tomczak, and Cheng Zhang (eds.), \emph{Advances in Neural Information Processing Systems 38: Annual Conference on Neural Information Processing Systems 2024, NeurIPS 2024, Vancouver, BC, Canada, December 10 - 15, 2024}, 2024.

\bibitem[Snoek et~al.(2018)Snoek, Wiltschko, and Rahimi]{sculley2018winner}
Jasper Snoek, Alex Wiltschko, and Ali Rahimi.
\newblock Winner's curse? on pace, progress, and empirical rigor.
\newblock 2018.

\bibitem[Soni et~al.(2024)Soni, Nathani, and Mishra]{10825393}
Rajat Soni, Mohit Nathani, and Rajiv Mishra.
\newblock Comprehensive evaluation of deep ltsf models for forecasting of air quality index.
\newblock In \emph{2024 IEEE International Conference on Big Data (BigData)}, pp.\  4410--4419, 2024.
\newblock \doi{10.1109/BigData62323.2024.10825393}.

\bibitem[Stine(2006)]{stine2006comment}
RA~Stine.
\newblock Comment: Classifier technology and the illusion of progress.
\newblock \emph{Statistical science}, \penalty0 (1):\penalty0 27--29, 2006.

\bibitem[Sun et~al.(2024)Sun, Li, Li, and Hong]{sun2023test}
Chenxi Sun, Hongyan Li, Yaliang Li, and Shenda Hong.
\newblock Test: Text prototype aligned embedding to activate {LLM}'s ability for time series.
\newblock In \emph{The Twelfth International Conference on Learning Representations}, 2024.

\bibitem[Tan et~al.(2024)Tan, Merrill, Gupta, Althoff, and Hartvigsen]{tan_are_2024}
Mingtian Tan, Mike~A. Merrill, Vinayak Gupta, Tim Althoff, and Thomas Hartvigsen.
\newblock Are {Language} {Models} {Actually} {Useful} for {Time} {Series} {Forecasting}?
\newblock \emph{Advances in Neural Information Processing Systems}, 37:\penalty0 60162--60191, December 2024.
\newblock URL \url{https://proceedings.neurips.cc/paper_files/paper/2024/hash/6ed5bf446f59e2c6646d23058c86424b-Abstract-Conference.html}.

\bibitem[Tan et~al.(2019)Tan, Chen, Pang, Vasudevan, Sandler, Howard, and Le]{tan2019mnasnet}
Mingxing Tan, Bo~Chen, Ruoming Pang, Vijay Vasudevan, Mark Sandler, Andrew Howard, and Quoc~V Le.
\newblock Mnasnet: Platform-aware neural architecture search for mobile.
\newblock In \emph{Proceedings of the IEEE/CVF conference on computer vision and pattern recognition}, pp.\  2820--2828, 2019.

\bibitem[Toda \& Phillips(1993)Toda and Phillips]{toda1993}
Hiro~Y. Toda and Peter C.~B. Phillips.
\newblock Vector autoregressions and causality.
\newblock \emph{Econometrica}, 61\penalty0 (6):\penalty0 1367--1393, 1993.
\newblock \doi{10.2307/2951647}.

\bibitem[Wang et~al.(2024{\natexlab{a}})Wang, Wu, Shi, Hu, Luo, Ma, Zhang, and ZHOU]{wang_timemixer_2024}
Shiyu Wang, Haixu Wu, Xiaoming Shi, Tengge Hu, Huakun Luo, Lintao Ma, James~Y. Zhang, and JUN ZHOU.
\newblock Timemixer: Decomposable multiscale mixing for time series forecasting.
\newblock In \emph{The Twelfth International Conference on Learning Representations}, 2024{\natexlab{a}}.

\bibitem[Wang et~al.(2024{\natexlab{b}})Wang, Wu, Dong, Liu, Long, and Wang]{wang_deep_2024}
Yuxuan Wang, Haixu Wu, Jiaxiang Dong, Yong Liu, Mingsheng Long, and Jianmin Wang.
\newblock Deep {Time} {Series} {Models}: {A} {Comprehensive} {Survey} and {Benchmark}, July 2024{\natexlab{b}}.
\newblock URL \url{http://arxiv.org/abs/2407.13278}.
\newblock arXiv:2407.13278 [cs].

\bibitem[Wang et~al.(2024{\natexlab{c}})Wang, Wu, Dong, Qin, Zhang, Liu, Qiu, Wang, and Long]{wang_timexer_2024}
Yuxuan Wang, Haixu Wu, Jiaxiang Dong, Guo Qin, Haoran Zhang, Yong Liu, Yunzhong Qiu, Jianmin Wang, and Mingsheng Long.
\newblock Timexer: Empowering transformers for time series forecasting with exogenous variables.
\newblock In \emph{The Thirty-eighth Annual Conference on Neural Information Processing Systems}, 2024{\natexlab{c}}.

\bibitem[Wang et~al.(2025)Wang, Kong, Feng, Wang, Yang, Zhao, Wang, and Zhang]{wang2025mamba}
Zihan Wang, Fanheng Kong, Shi Feng, Ming Wang, Xiaocui Yang, Han Zhao, Daling Wang, and Yifei Zhang.
\newblock Is mamba effective for time series forecasting?
\newblock \emph{Neurocomputing}, 619:\penalty0 129178, 2025.

\bibitem[Weron(2014)]{WERON20141030}
Rafał Weron.
\newblock Electricity price forecasting: A review of the state-of-the-art with a look into the future.
\newblock \emph{International Journal of Forecasting}, 30\penalty0 (4):\penalty0 1030--1081, 2014.
\newblock ISSN 0169-2070.
\newblock \doi{https://doi.org/10.1016/j.ijforecast.2014.08.008}.
\newblock URL \url{https://www.sciencedirect.com/science/article/pii/S0169207014001083}.

\bibitem[Woo et~al.(2024)Woo, Liu, Kumar, Xiong, Savarese, and Sahoo]{woo2024unified}
Gerald Woo, Chenghao Liu, Akshat Kumar, Caiming Xiong, Silvio Savarese, and Doyen Sahoo.
\newblock Unified training of universal time series forecasting transformers.
\newblock In \emph{Forty-first International Conference on Machine Learning}, 2024.

\bibitem[Wu et~al.(2021)Wu, Xu, Wang, and Long]{wu2021autoformer}
Haixu Wu, Jiehui Xu, Jianmin Wang, and Mingsheng Long.
\newblock Autoformer: Decomposition transformers with auto-correlation for long-term series forecasting.
\newblock \emph{Advances in neural information processing systems}, 34:\penalty0 22419--22430, 2021.

\bibitem[Wu et~al.(2023)Wu, Hu, Liu, Zhou, Wang, and Long]{wu_timesnet_2023}
Haixu Wu, Tengge Hu, Yong Liu, Hang Zhou, Jianmin Wang, and Mingsheng Long.
\newblock Timesnet: Temporal 2d-variation modeling for general time series analysis.
\newblock In \emph{The Eleventh International Conference on Learning Representations}, 2023.

\bibitem[Wu \& Keogh(2023)Wu and Keogh]{wu_current_2023}
Renjie Wu and Eamonn~J. Keogh.
\newblock Current {Time} {Series} {Anomaly} {Detection} {Benchmarks} are {Flawed} and are {Creating} the {Illusion} of {Progress}.
\newblock \emph{IEEE Transactions on Knowledge and Data Engineering}, 35\penalty0 (3):\penalty0 2421--2429, March 2023.
\newblock ISSN 1558-2191.
\newblock \doi{10.1109/TKDE.2021.3112126}.
\newblock URL \url{https://ieeexplore.ieee.org/document/9537291}.

\bibitem[Xu et~al.(2025)Xu, Gupta, Cheng, Shen, Shen, Talwalkar, and Khodak]{xu_specialized_2024}
Zongzhe Xu, Ritvik Gupta, Wenduo Cheng, Alexander Shen, Junhong Shen, Ameet Talwalkar, and Mikhail Khodak.
\newblock Specialized foundation models struggle to beat supervised baselines.
\newblock In \emph{The Thirteenth International Conference on Learning Representations, {ICLR} 2025, Singapore, April 24-28, 2025}, 2025.

\bibitem[Xue \& Salim(2023)Xue and Salim]{xue2023promptcast}
Hao Xue and Flora~D Salim.
\newblock Promptcast: A new prompt-based learning paradigm for time series forecasting.
\newblock \emph{IEEE Transactions on Knowledge and Data Engineering}, 2023.

\bibitem[Yao et~al.(2025)Yao, Yang, Jiang, Liang, Jin, and Pan]{yao2025towards}
Qingren Yao, Chao-Han~Huck Yang, Renhe Jiang, Yuxuan Liang, Ming Jin, and Shirui Pan.
\newblock Towards neural scaling laws for time series foundation models.
\newblock In \emph{The Thirteenth International Conference on Learning Representations}, 2025.

\bibitem[Zeng et~al.(2023)Zeng, Chen, Zhang, and Xu]{zeng_are_2022}
Ailing Zeng, Muxi Chen, Lei Zhang, and Qiang Xu.
\newblock Are transformers effective for time series forecasting?
\newblock In \emph{The Thirty-Seventh AAAI Conference on Artificial Intelligence (AAAI-23)}. AAAI Press, 2023.
\newblock ISBN 978-1-57735-880-0.
\newblock \doi{10.1609/aaai.v37i9.26317}.
\newblock URL \url{https://doi.org/10.1609/aaai.v37i9.26317}.

\bibitem[Zhang \& Yan(2023)Zhang and Yan]{zhang2023crossformer}
Yunhao Zhang and Junchi Yan.
\newblock Crossformer: Transformer utilizing cross-dimension dependency for multivariate time series forecasting.
\newblock In \emph{The eleventh international conference on learning representations}, 2023.

\bibitem[Zhao et~al.(2025)Zhao, Shen, Liu, Wang, and Deng]{less-is-more-prune-then-finetune}
Lifan Zhao, Yanyan Shen, Zhaoyang Liu, Xue Wang, and Jiaji Deng.
\newblock Less is more: Unlocking specialization of time series foundation models via structured pruning.
\newblock In \emph{The Thirty-ninth Annual Conference on Neural Information Processing Systems}, 2025.
\newblock URL \url{https://openreview.net/forum?id=jy4bBsr1Jc}.

\bibitem[Zhou et~al.(2021)Zhou, Zhang, Peng, Zhang, Li, Xiong, and Zhang]{zhou2021informer}
Haoyi Zhou, Shanghang Zhang, Jieqi Peng, Shuai Zhang, Jianxin Li, Hui Xiong, and Wancai Zhang.
\newblock Informer: Beyond efficient transformer for long sequence time-series forecasting.
\newblock In \emph{Proceedings of the AAAI conference on artificial intelligence}, volume~35, pp.\  11106--11115, 2021.

\bibitem[Zhou et~al.(2022)Zhou, Ma, Wen, Wang, Sun, and Jin]{zhou2022fedformer}
Tian Zhou, Ziqing Ma, Qingsong Wen, Xue Wang, Liang Sun, and Rong Jin.
\newblock Fedformer: Frequency enhanced decomposed transformer for long-term series forecasting.
\newblock In \emph{International conference on machine learning}, pp.\  27268--27286. PMLR, 2022.

\end{thebibliography}
